\documentclass{article}
\PassOptionsToPackage{numbers, compress}{natbib}

\usepackage[preprint]{neurips_2026}

\usepackage[utf8]{inputenc}
\usepackage[T1]{fontenc}
\usepackage{hyperref}
\usepackage{url}
\usepackage{booktabs}
\usepackage{amsfonts}
\usepackage{amsmath}
\usepackage{nicefrac}
\usepackage{microtype}
\usepackage{xcolor}
\usepackage{graphicx}
\usepackage{multirow}
\usepackage{enumitem}
\usepackage{subcaption}
\usepackage{wrapfig}

\title{Encoder-Free Human Motion Understanding via Structured Motion Descriptions}

\author{%
  Yao Zhang \\
  Aalto University, Espoo, Finland \\
  \texttt{yao.1.zhang@aalto.fi} \\
  \And
  Zhuchenyang Liu \\
  Aalto University, Espoo, Finland \\
  \texttt{zhuchenyang.liu@aalto.fi} \\
  \AND
  Thomas Ploetz \\
  Georgia Institute of Technology, Atlanta, GA, USA \\
  \texttt{thomas.ploetz@gatech.edu} \\
  \And
  Yu Xiao \\
  Aalto University, Espoo, Finland \\
  \texttt{yu.xiao@aalto.fi} \\
}

\begin{document}

\maketitle

\begin{abstract}
The world knowledge and reasoning capabilities of text-based large language models (LLMs) are advancing rapidly, yet current approaches to human motion understanding, including motion question answering and captioning, have not fully exploited these capabilities. Existing LLM-based methods typically learn motion-language alignment through dedicated encoders that project motion features into the LLM's embedding space, remaining constrained by cross-modal representation and alignment. Inspired by biomechanical analysis, where joint angles and body-part kinematics have long served as a precise descriptive language for human movement, we propose \textbf{Structured Motion Description (SMD)}, a rule-based, deterministic approach that converts joint position sequences into structured natural language descriptions of joint angles, body part movements, and global trajectory. By representing motion as text, SMD enables LLMs to apply their pretrained knowledge of body parts, spatial directions, and movement semantics directly to motion reasoning, without requiring learned encoders or alignment modules. We show that this approach goes beyond state-of-the-art results on both motion question answering (66.7\% on BABEL-QA, 90.1\% on HuMMan-QA) and motion captioning (R@1 of 0.584, CIDEr of 53.16 on HumanML3D), surpassing all prior methods. SMD additionally offers practical benefits: the same text input works across different LLMs with only lightweight LoRA adaptation (validated on 8 LLMs from 6 model families), and its human-readable representation enables interpretable attention analysis over motion descriptions. Code, data, and pretrained LoRA adapters are available at \url{https://yaozhang182.github.io/motion-smd/}.

\end{abstract}

\section{Introduction}
\label{sec:intro}

\begin{figure}[htbp]
\centering
\includegraphics[width=0.8\textwidth]{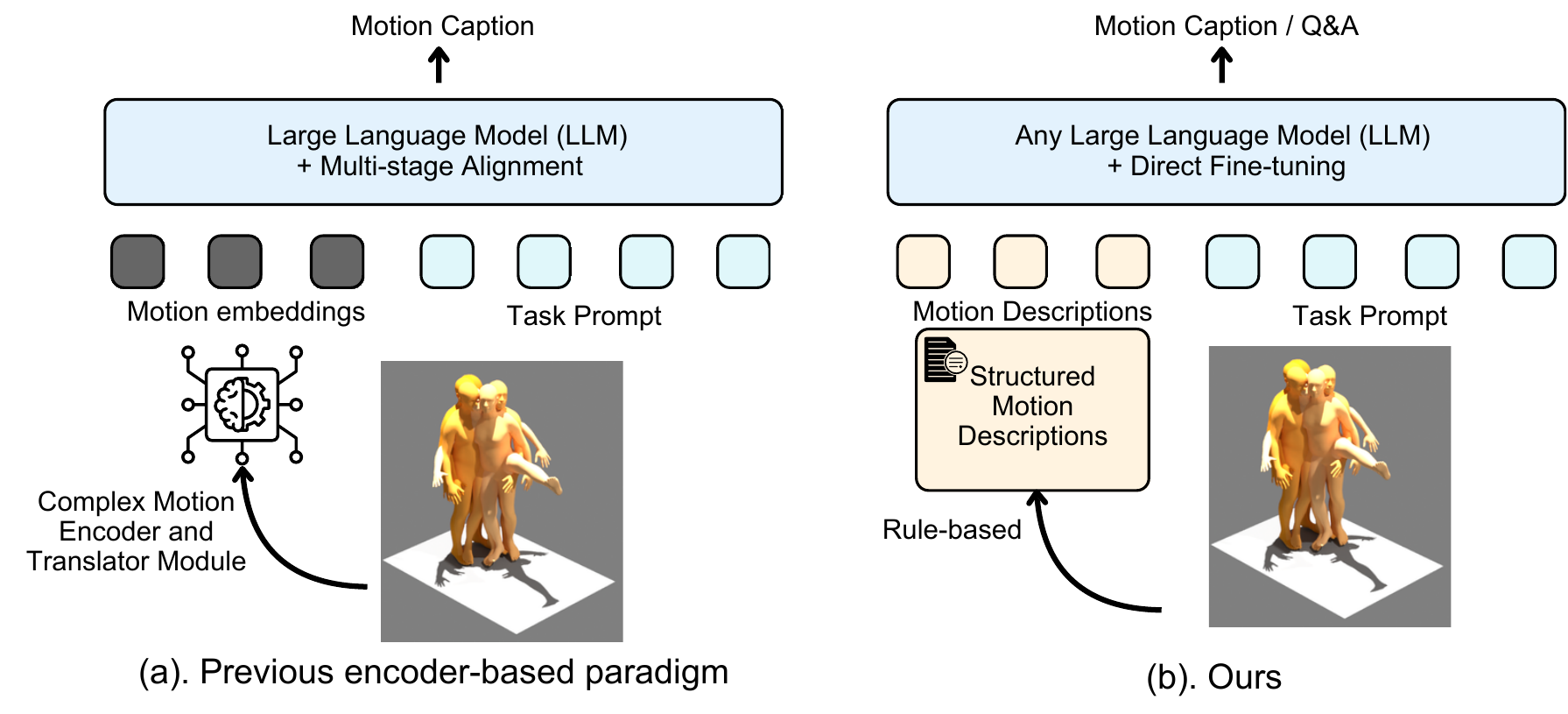}
\caption{Comparison of our encoder-free approach (b) against prior encoder-based methods (a).
}
\label{fig:pipeline}
\end{figure}

Understanding human motion from skeletal data, typically represented as sequences of 2D or 3D body joints connected by bones over time, is a fundamental problem in computer vision, underpinning tasks such as motion question answering (QA) and motion captioning.
Motion QA~\citep{endo2023humanmotionqa, liu2024imore} entails answering questions about a given motion sequence, such as which body part is moving, in which direction, or what action is being performed, which requires fine-grained spatiotemporal reasoning over joint-level motion details.
Motion captioning~\citep{guo2022t2m, chen2024motiongpt3} requires generating a natural language description that summarizes the overall motion content, bridging the gap between low-level skeletal data and high-level semantic understanding. The rapid advancement of large language models (LLMs) has motivated recent efforts to tackle these tasks with LLMs, leveraging their world knowledge and reasoning capabilities for natural language interaction with motion data.

Most of these LLM-based approaches ~\citep{chen2024motiongpt3, wu2025mgmotionllm, jiang2024motiongpt, chen2024motionllm, wang2024motiongpt2} follow an encoder-based paradigm: a learned motion encoder projects sequences of joint positions into the LLM's token embedding space, using either discrete tokenization via VQ-VAE~\citep{jiang2024motiongpt, wu2025mgmotionllm}, continuous latent encoding via VAE~\citep{chen2024motiongpt3}, or linear projection~\citep{chen2024motionllm}, following the multi-modal alignment framework established in vision-language models (VLMs) ~\citep{liu2024llava}.
These methods have driven substantial progress, yet the reliance on cross-modal representation and alignment introduces practical considerations: the motion encoder and alignment module are typically trained through multi-stage pipelines with paired motion-text data, the resulting system is coupled to a specific LLM backbone, and the learned motion tokens are not human-readable.

As illustrated in Fig.~\ref{fig:pipeline}, we take an encoder-free approach by introducing Structured Motion Description (SMD): a deterministic, rule-based conversion from joint-position sequences into structured natural-language text, which is then fed directly to LLMs. This idea leverages the fact that LLMs, pretrained on vast corpora, already encode extensive knowledge of body parts, spatial directions, and movement semantics~\citep{li2025llmmotion}. By replacing learned motion embeddings with descriptive text, SMD enables LLMs to process motion in their native modality.

The key challenge is to produce descriptions that are precise enough for fine-grained reasoning while using concepts the LLM already understands. Biomechanics provides a natural foundation: joint angles and body-part kinematics have long served as a precise descriptive language for human movement~\citep{perry2024gait, chen2024languageofmotion, zhang2026jami}. Clinical gait analysis, for example, characterizes walking via time-varying flexion curves of the hip, knee, and ankle. These descriptors are inherently textual; a statement like "hip flexion increases from 3° to 81°" is simultaneously a quantitative measurement and a natural-language sentence. SMD extends this idea to full-body motion by computing angular changes across all major joints together with global trajectory information.

Because SMD uses natural language terms for body parts, spatial directions, and temporal patterns, it naturally leverages the LLM's world knowledge of these concepts~\citep{li2025llmmotion}, as validated by our zero-shot experiments in Section~\ref{sec:exp}.
With lightweight LoRA~\citep{hu2022lora} fine-tuning as the only training step, our approach delivers three key advantages:
(1) \textbf{Strong performance with minimal training}: SMD surpasses prior state-of-the-art on motion QA~\citep{endo2023humanmotionqa, liu2024imore} and captioning~\citep{guo2022t2m} without any learned motion encoder or multimodal alignment module, outperforming both dedicated motion models and encoder-based LLM approaches;
(2) \textbf{LLM-agnostic flexibility at low cost}: the same text input works across different LLMs by attaching a lightweight LoRA adapter ($\sim$40M parameters, 2--8 GPU-hours on a single H200), without retraining a motion encoder or alignment module; we validate this across 8 LLMs from 6 model families;
(3) \textbf{Built-in interpretability}: attention analysis over the human-readable SMD tokens directly reveals which body parts and trajectory segments the model relies on for generation, offering transparent, fine-grained interpretability by design.

\section{Related Work}
\label{sec:related}

\paragraph{Motion representation.}
3D human motion is most commonly represented as joint-position sequences from parametric body models such as SMPL~\citep{loper2015smpl}, typically via the 263-dim HumanML3D feature~\citep{guo2022t2m} (root velocities, local joint positions, 6D joint rotations~\citep{zhou2019continuity}, joint velocities, and foot contact labels); while expressive, these high-dimensional numerical features require learned encoders to interface with LLMs.
A complementary line uses interpretable, human-readable descriptors: PoseScript~\citep{delmas2022posescript} and PoseFix~\citep{delmas2023posefix} generate rule-based natural-language descriptions of static poses based on joint positions and predefined rules for retrieval and correction tasks, while \citet{zhang2026jami} map biomechanical joint angles from skeleton sequences into structured pseudo-images for motion retrieval with vision transformers. Our work also builds on joint-angle representations, but converts them into structured text rather than images, enabling any LLM to process motion directly without learned visual encoders.

\paragraph{Motion understanding.}
Human motion understanding encompasses a range of tasks that reason over skeletal sequences. We focus on two representative tasks: motion QA and motion captioning.
For motion QA, early methods use task-specific architectures without pretrained LLMs: NSPose~\citep{endo2023humanmotionqa} introduces the BABEL-QA benchmark with a neuro-symbolic framework that parses questions into compositional programs executed by hand-designed modules over learned motion features, while IMoRe~\citep{liu2024imore} replaces these modules with implicit program-guided reasoning. MotionLLM~\citep{chen2024motionllm} was the first to apply a billion-parameter LLM (Vicuna-13B) via a learned motion encoder, achieving performance comparable to IMoRe.

For motion captioning, early approaches build on motion-text joint embedding spaces (TM2T~\citep{guo2022tm2t}, LaMP~\citep{li2024lampm2t}, MoTe~\citep{wu2024mote}). With the rise of LLMs, recent methods have shifted toward the encoder-alignment paradigm established by VLMs such as LLaVA~\citep{liu2024llava}. MotionGPT~\citep{jiang2024motiongpt} uses VQ-VAE tokens on GPT, MotionGPT-2~\citep{wang2024motiongpt2} extends this to LLaMA-3.1-8B, MoChat~\citep{mo2024mochat} adds multi-turn grounding, MotionGPT3~\citep{chen2024motiongpt3} replaces discrete tokens with a continuous VAE latent trained in three stages on GPT-2, and MG-MotionLLM~\citep{wu2025mgmotionllm} trains a multi-granularity framework on T5 across 28 motion-language tasks. All share learned motion encoders and alignment modules coupled to a specific LLM backbone. Complementing this line, \citet{li2025llmmotion} find that pretrained LLMs already possess knowledge of body parts and physics but need substantial adaptation for precise motion tasks. Beyond motion, text-mediated LLM inputs have been explored in video understanding by works such as LLoVi~\citep{zhang2024llovi} and Socratic Models~\citep{zeng2022socratic}, but these rely on a learned caption model (e.g., a VLM) to produce the text, inheriting its biases and losing fine-grained spatial-temporal detail. Skeletal sequences instead have a structured, low-dimensional form that admits precise, deterministic conversion to text without any learned component. Our work converts motion into structured biomechanical descriptions, enabling any LLM to perform motion understanding without learned encoders or cross-modal alignment.

\section{Method}
\label{sec:method}

\begin{figure}[htbp]
\centering
\includegraphics[width=\textwidth]{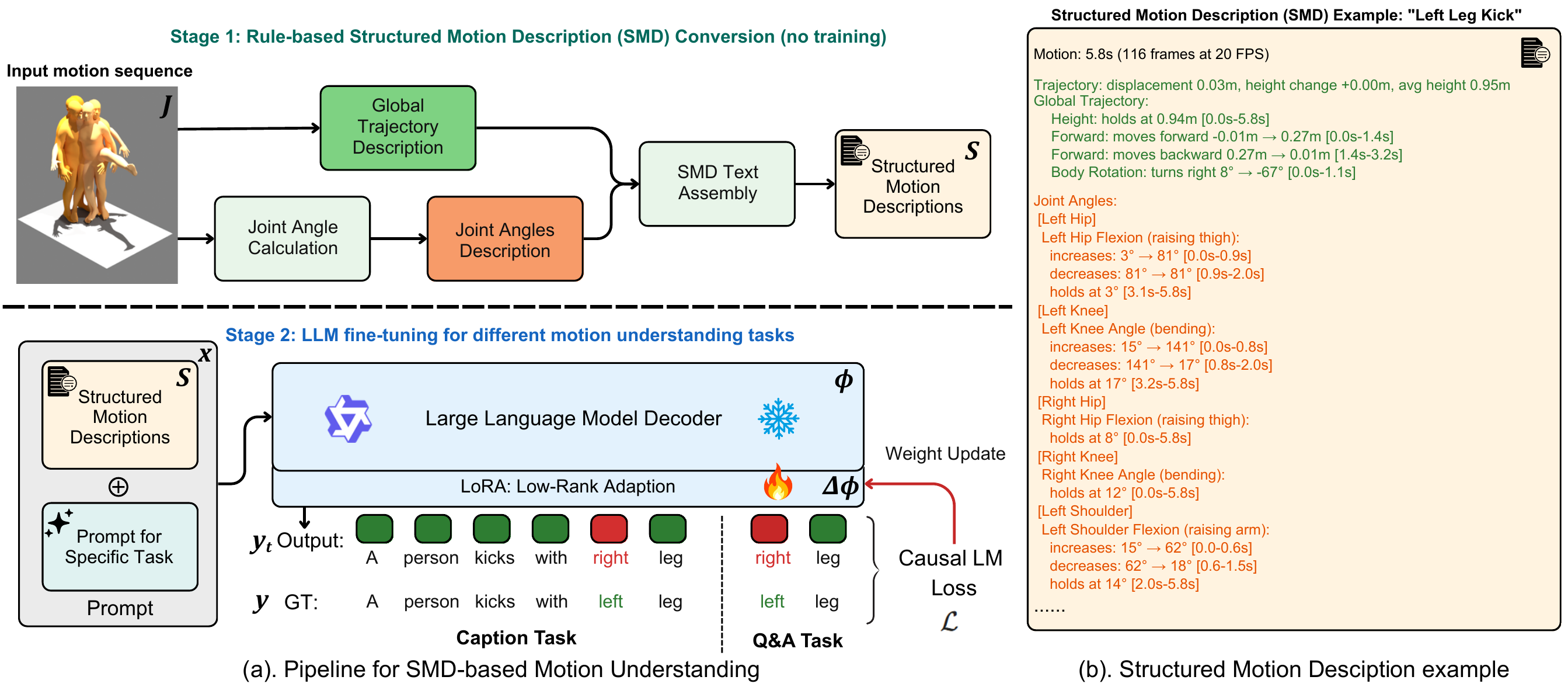}
\caption{Overview of our approach.(a) Stage 1: conversion of skeleton sequences into SMD; Stage 2: SMD fed into an LLM fine-tuned with LoRA.
(b) Truncated SMD for a ``Left Leg Kick'' motion, organized into a meta-information header, a global trajectory block, and a joint angles block grouped by body part. Only the three most active joints are shown; the full 26-angle SMD is in Appendix~\ref{app:angles}.}
\label{fig:method}
\end{figure}

As illustrated in Figure~\ref{fig:method}, our approach consists of two stages: (1) a rule-based conversion $f_\text{SMD}$ from joint positions to an SMD, and (2) LoRA fine-tuning of a pretrained LLM that takes this text as input.
No learned motion encoder, VQ-VAE, or cross-modal alignment module is involved.

\subsection{Structured Motion Description}
\label{sec:smd}
$f_\text{SMD}: \mathbb{R}^{T \times J \times 3} \to \mathcal{V}^*$ is a deterministic function that maps a sequence of joint positions  $\mathbf{J} = [\mathbf{j}_1, \dots, \mathbf{j}_T] \in \mathbb{R}^{T \times J \times 3}$ (where $J=22$ for SMPL-based representation and $T$ is the number of frames) to a text string $\mathcal{S} \in \mathcal{V}^*$ over the LLM's vocabulary $\mathcal{V}$. As illustrated in Figure~\ref{fig:method}(a), the conversion proceeds in the following four steps.

\paragraph{Step 1: Joint angle calculation.}
The goal of this step is to produce a set of $K$ joint-angle channels $\{\theta_k^{(1:T)}\}_{k=1}^{K}$, each a scalar time series of length $T$ that describes the angle at one anatomical joint over the input frames. Following standard biomechanical conventions~\citep{zhang2026jami, schlegel2024joint, wu2002isb, wu2005isb}, we define these angles by projecting bone vectors onto anatomical reference planes within joint-local coordinate frames, organized along a kinematic chain rooted at a body-local coordinate frame.

The body-local coordinate frame is a right-handed reference system attached to the pelvis, with $\mathbf{e}_x$ pointing forward, $\mathbf{e}_y$ pointing up, and $\mathbf{e}_z$ pointing laterally (right). It defines the subject's overall facing direction and serves as the root of the kinematic chain. At each time frame $t$, we construct it from three landmark joints (pelvis, left hip, right hip), and then compute joint angles sequentially along the chain. For example, hip flexion is measured in the sagittal ($\mathbf{e}_x$--$\mathbf{e}_y$) plane as the signed angle of the femur vector $\mathbf{v}_\text{fem} = \mathbf{j}_\text{knee} - \mathbf{j}_\text{hip}$ from the downward direction $-\mathbf{e}_y$ toward $\mathbf{e}_x$:
\begin{equation}
\theta_\text{hip-flex}^{(t)} = \mathrm{atan2}\!\left(\hat{\mathbf{v}}_\text{fem}^{(t)} \cdot \mathbf{e}_x,\; -\hat{\mathbf{v}}_\text{fem}^{(t)} \cdot \mathbf{e}_y\right).
\label{eq:hip_flex}
\end{equation}
This yields a signed angle in [-180\textdegree, 180\textdegree], so flexion (positive) and extension (negative) are distinguished naturally. Adduction is computed as the angle between the femur vector and its projection onto the sagittal plane, and rotation about the femur is computed using the tibia direction as a secondary reference.
Other angles, such as knee flexion and shoulder adduction, follow analogous definitions using appropriate bone vectors and reference planes ( Appendix~\ref{app:angles} for the complete list).

In total, this procedure yields $K = 26$ biomechanical joint angles $\boldsymbol{\theta}^{(t)} = [\theta_1^{(t)}, \dots, \theta_K^{(t)}]$, organized into 13 body-part groups spanning the pelvis, lumbar spine, neck, bilateral hip, knee, ankle, shoulder, and elbow. Because these angles follow standard anatomical conventions rather than skeleton-specific indices, SMD generalizes to other skeleton formats (e.g., Kinect, COCO) provided the relevant landmarks are available, as we demonstrate on NTU RGB+D~60 in Appendix~\ref{app:cross_skeleton}; each angle also carries a clear anatomical meaning (e.g., hip flexion = raising the thigh, shoulder adduction = moving the arm inward), enabling a direct translation into text.

\paragraph{Step 2: Global trajectory description.}
Joint angles describe how body parts are posed relative to one another, but they are invariant to the subject's global position and orientation in the world. Many motions (e.g., walking, jumping, turning), however, are also defined by global displacement of the body through space. We characterize global motion using the pelvis position, since it is the root joint of the SMPL skeleton. The pelvis also closely approximates the body's center of mass, making it a standard landmark for global motion analysis in biomechanics and gait research~\citep{perry2024gait}.

Specifically, we extract the pelvis trajectory along three translational axes (forward/backward, lateral, height) and one rotational axis (body yaw, computed from the forward direction $\mathbf{e}_x$ established in Step 1). Each axis is segmented into contiguous intervals using a two-stage procedure: the time series is first smoothed with a moving average filter of window size $w = 7$ frames (0.35s at 20 FPS, short enough to preserve rapid voluntary movements while filtering frame-level noise), and then partitioned by peak-valley detection whenever the change between consecutive extrema exceeds a minimum threshold $\tau_p$ ($0.03$m for translation, to ignore positional jitter; $15$\textdegree{} for yaw, to distinguish deliberate turns from postural sway). Each resulting segment is classified according to the sign and axis of motion, yielding descriptions such as ``Forward Position: moves forward 0.00m $\to$ 1.23m [0.0s--2.5s]'' or ``Body Rotation: turns left 0\textdegree{} $\to$ 45\textdegree{} [1.0s--1.8s]'' (the complete set of descriptive phrases is listed in Table~\ref{tab:smd_vocab}, Appendix~\ref{app:angles}). We denote the resulting set of trajectory segments as $\{s_\text{traj}\}$. A sensitivity analysis of the Segmentation thresholds $\tau_p$ and Smoothing window $w$ is provided in Table~\ref{tab:smd_sensitivity}.

\paragraph{Step 3: Joint angles description.}
We apply the same smoothing and peak-valley segmentation procedure as in Step 2 to each of the $K$ time series $\theta_k^{(1:T)}$ from Step 1, using a minimum angular change threshold of $\delta = 5$\textdegree{} (set above the noise floor of the input data to avoid spurious segments; sensitivity analysis in Table~\ref{tab:smd_sensitivity}). Each segment is labeled with one of four types based on the boundary values $v_\text{start}$ and $v_\text{end}$ of the segment: \emph{increases} if the angle rises by at least $\delta$; \emph{decreases} if it falls by at least $\delta$; \emph{holds} if the change is smaller than $\delta$; and \emph{repeats $N$ cycles} if consecutive moving segments alternate in direction with consistent amplitude (Appendix~\ref{app:angles}), capturing periodic patterns such as cyclic hip and knee flexion during walking.

As a concrete example, a left hip flexion angle that rises from 3\textdegree{} to 81\textdegree{} over the first 18 frames, falls back to 7\textdegree{} by frame 40, and then stays near 3\textdegree{} is represented as three segments: "Left Hip Flexion (raising thigh): \emph{increases} $3\text{\textdegree} \to 81\text{\textdegree}$ [0.0s--0.9s], \emph{decreases} $81\text{\textdegree} \to 7\text{\textdegree}$ [0.9s--2.0s], \emph{holds at} $3\text{\textdegree}$ [2.0s--5.8s]". This compression reduces a raw time series of $T$ values (e.g., 200 frames) into typically 3--8 descriptive intervals per joint, substantially shortening the text while preserving the temporal structure. We denote the resulting text segments as $\{s_k\}_{k=1}^{K}$, one group per joint angle.

\paragraph{Step 4: SMD assembly.}
The trajectory and joint angle segments from Steps 2 and 3 are assembled into a single structured text string $\mathcal{S} = f_\text{assemble}(\{s_k\}_{k=1}^K, \{s_\text{traj}\})$. As shown in the ``Left Leg Kick'' example in Figure~\ref{fig:method}(b), $\mathcal{S}$ is organized hierarchically into three blocks:
\begin{itemize}
\item \textbf{Meta information}: a one-line header reporting motion duration, frame count, and frame rate (e.g., ``Motion: 5.8s (116 frames at 20 FPS)'').
\item \textbf{Global trajectory}: an aggregated summary (overall displacement, height change, average height) followed by per-axis trajectory segments from Step 2 (e.g., ``Forward: moves forward -0.01m $\to$ 0.27m [0.0s--1.4s]''; ``Body Rotation: turns right 8\textdegree{} $\to$ -67\textdegree{} [0.0s--1.1s]'').
\item \textbf{Joint angles}: per-joint angle segments from Step 3, grouped by body part with section headers (e.g., ``[Left Hip]'', ``[Left Knee]'').
\end{itemize}
Once tokenized by the LLM, $\mathcal{S}$ occupies approximately 4{,}000 tokens on HumanML3D for all 26 joint angles, or approximately 1{,}000 input tokens when only the Top-3 most active joints are selected.

\subsection{Task Formulation and Training}
\label{sec:finetuning}

We formulate both motion QA and motion captioning tasks as autoregressive text generation conditioned on the SMD. Let $\text{LLM}_\phi$ denote a pretrained LLM with frozen parameters $\phi$. Given a motion $\mathbf{J}$, we compute $\mathcal{S} = f_\text{SMD}(\mathbf{J})$ and construct the text prompt ${x}$ by concatenating a task-specific system instruction with $\mathcal{S}$; for QA we additionally append the question text $q$ and the candidate answer options, and the target ${y} = [y_1, \dots, y_L]$ is the text of the correct option, whereas for captioning the target ${y}$ is a natural-language caption describing the motion. The multiple-choice QA format matches the standard protocol of existing motion QA benchmarks~\citep{endo2023humanmotionqa, liu2024imore} and the evaluation setting of all baselines; it naturally extends to open-ended QA by omitting the options. The full prompt template for both tasks is shown in Appendix~\ref{app:prompt_format}.

During training, only the target response tokens ${y}$ contribute to the loss; all tokens in ${x}$ are masked.
The training objective is the standard causal language modeling loss over ${y}$:
\begin{equation}
\mathcal{L} = -\sum_{l=1}^{L} \log p_{\phi + \Delta\phi}(y_l \mid \mathbf{x}, y_{1:l-1}),
\label{eq:loss}
\end{equation}
where $\Delta\phi$ are LoRA~\citep{hu2022lora} adapter parameters inserted into all linear layers, with the base LLM weights $\phi$ frozen. We train separate LoRA adapters for motion QA and motion captioning.

At inference, given a new motion $\mathbf{J}'$, we build the prompt $\mathbf{x}'$ from $\mathcal{S}' = f_\text{SMD}(\mathbf{J}')$ (without the target) and generate $\hat{\mathbf{y}} = \text{LLM}_{\phi + \Delta\phi}(\mathbf{x}')$ autoregressively; QA predictions are compared to the ground-truth answer, and captions are evaluated against references with standard metrics.

\section{Experiments and Analysis}
\label{sec:exp}

\subsection{Experimental Setup}

\paragraph{Datasets.}
We evaluate on BABEL-QA~\citep{endo2023humanmotionqa} (1,109 motions, 2,577 QA pairs, 1,800/384/393 train/val/test) and HuMMan-QA~\citep{liu2024imore} (925 motions, 3,123 QA pairs, 2,066/524/533 train/val/test) for motion QA, and HumanML3D~\citep{guo2022t2m} (14,616 motions with 44,970 captions) for motion captioning. Since the original QA benchmarks use variable option counts (4–20 for BABEL-QA, 6–155 for HuMMan-QA), making cross-method comparison difficult, we standardize both to a fixed 10-option format (details in Appendix~\ref{app:experiments_setup}). All methods are retrained and evaluated on this format except MotionLLM~\citep{chen2024motionllm}, whose training data and checkpoints are unavailable (a fair comparison in Appendix~\ref{app:openvocab}).

\paragraph{Evaluation metrics.}
For QA, we report accuracy via exact string matching following previous works~\citep{chen2024motionllm, liu2024imore, endo2023humanmotionqa}. For captioning, we use text-motion alignment metrics (R-Precision R@1/2/3 and MM-Distance computed with the pretrained T2M evaluator~\citep{guo2022t2m}) and linguistic metrics (BLEU@1/4~\citep{papineni2002bleu}, ROUGE-L~\citep{lin2004rouge}, CIDEr~\citep{vedantam2015cider}, BERTScore~\citep{zhang2019bertscore}) following the common evaluation protocol used in prior works~\citep{chen2024motiongpt3, wang2024motiongpt2, wu2025mgmotionllm, jiang2024motiongpt, guo2022tm2t}. Since prior CIDEr computations inadvertently included extraneous symbols (e.g., colons, newlines) affecting scores, we re-evaluate CIDEr for all baselines on their released checkpoints with a cleaned implementation; details in Appendix~\ref{app:metrics}.

\paragraph{Implementation details.}
Our default backbone is Qwen2.5-7B-Instruct.
LoRA (rank 16, $\alpha = 32$, dropout 0.05) is applied to all linear layers, yielding approximately 40M trainable parameters out of 7.6B total; the base model weights remain frozen.
For QA, we train jointly on BABEL-QA (1{,}800 pairs) and HuMMan-QA (2{,}066 pairs) for 5 epochs with batch size 8, learning rate 1e-4, and cosine annealing (approximately 7 GPU-hours).
For captioning, we train on HumanML3D train set (23{,}384 motions, each with 3--4 reference captions) for 5 epochs with batch size 8 and learning rate 1e-4 (approximately 20 GPU-hours).
All experiments use a single NVIDIA H200 GPU.


\paragraph{Baselines.}
For QA, we compare against specialized motion QA models (NSPose~\citep{endo2023humanmotionqa}, IMoRe~\citep{liu2024imore}), an encoder-based motion LLM (MotionLLM~\citep{chen2024motionllm}), and MotionGPT3-Qwen. All QA baselines except MotionLLM~\citep{chen2024motionllm} are retrained on the standardized 10-option format for fair comparison. For captioning, we compare against TM2T~\citep{guo2022tm2t}, MotionGPT~\citep{jiang2024motiongpt}, LaMP~\citep{li2024lampm2t}, MoTe~\citep{wu2024mote}, MotionGPT3~\citep{chen2024motiongpt3}, MG-MotionLLM~\citep{wu2025mgmotionllm}, and MotionGPT3-Qwen. MotionGPT3-Qwen is a controlled baseline we construct by porting the MotionGPT3 paradigm (frozen VAE + MLP projection, LLaVA-style two-stage training) to our Qwen2.5-7B backbone, isolating the effect of motion representation while holding the LLM constant (construction details in Appendix~\ref{app:experiments_setup}).

\subsection{Main Results}

\begin{table}[tb]
\centering
\caption{Main results. Top: motion QA accuracy (\%) on BABEL-QA and HuMMan-QA. Bottom: motion captioning performance on the HumanML3D test set. $^\dagger$Results from the original paper. $^\ddagger$MotionGPT3 paradigm (pretrained VAE + MLP projection) replicated on the same Qwen2.5-7B backbone with LLaVA-style two-stage training (4 projection tokens).}
\label{tab:main}
\resizebox{0.7\textwidth}{!}{%
\begin{tabular}{llccc}
\toprule
\multicolumn{5}{c}{\textsc{Motion QA}} \\
\midrule
Method & Input & Model & BABEL-QA & HuMMan-QA \\
\midrule
NSPose~\citep{endo2023humanmotionqa} & joints & Specialized & 48.1 & 70.9 \\
IMoRe~\citep{liu2024imore} & joints & Specialized & 60.1 & 75.2 \\
MotionLLM$^\dagger$~\citep{chen2024motionllm} & joints & Vicuna-13B & 43.6 & --- \\
MotionGPT3-Qwen$^\ddagger$ & joints & Qwen2.5-7B & 50.1 & 22.0 \\
\textbf{Ours (SMD-based)} & text & Qwen2.5-7B & \textbf{66.7} & \textbf{90.1} \\
\bottomrule
\end{tabular}%
}

\resizebox{0.9\textwidth}{!}{%
\begin{tabular}{lccccccccc}
\toprule
\multicolumn{10}{c}{\textsc{Motion Captioning (M2T)}} \\
\midrule
Method & R@1$\uparrow$ & R@2$\uparrow$ & R@3$\uparrow$ & MMD$\downarrow$ & B@1$\uparrow$ & B@4$\uparrow$ & R-L$\uparrow$ & CIDEr$\uparrow$ & BS$\uparrow$ \\
\midrule
TM2T~\citep{guo2022tm2t} & .516 & --- & .823 & 2.94 & 48.9 & 7.0 & 38.1 & 35.1 & 32.2 \\
MotionGPT~\citep{jiang2024motiongpt} & .543 & --- & .827 & 2.82 & 48.2 & 12.5 & 37.4 & 39.4 & 32.4 \\
LaMP~\citep{li2024lampm2t} & .547 & --- & .831 & 2.81 & 47.8 & 13.0 & 37.1 & 39.7 & 32.7 \\
MoTe~\citep{wu2024mote} & .577 & --- & .871 & 2.65 & 46.7 & 11.2 & 37.4 & 41.3 & 30.3 \\
MG-MotionLLM~\citep{wu2025mgmotionllm} & \textbf{.592} & --- & .866 & 2.58 & --- & 8.1 & --- & --- & 36.7 \\
MotionGPT3~\citep{chen2024motiongpt3} & .573 & .773 & .864 & 2.43 & 59.1 & 19.4 & 46.2 & 40.6 & 35.2 \\
MotionGPT3-Qwen$^\ddagger$ & .555 & .753 & .843 & 2.47 & 59.43 & 19.60 & 45.82 & 46.13 & 42.87 \\
\textbf{Ours (SMD-based)} & .584 & \textbf{.794} & \textbf{.883} & \textbf{2.35} & \textbf{63.45} & \textbf{22.67} & \textbf{47.80} & \textbf{53.16} & \textbf{45.58} \\
\bottomrule
\end{tabular}%
}
\end{table}

\paragraph{Motion QA.}
As listed in Table \ref{tab:main}, our approach achieves 66.7\% accuracy on BABEL-QA and 90.1\% on HuMMan-QA, surpassing the previous state-of-the-art, IMoRe~\citep{liu2024imore}, by 6.6 and 14.9 percentage points, respectively.
On BABEL-QA, MotionGPT3-Qwen, which uses the same Qwen2.5-7B backbone but replaces SMD with VAE-encoded latent tokens, achieves 50.1\% accuracy, trailing our approach by 16.6 percentage points despite identical LLM capacity.
On HuMMan-QA, the gap is even larger: 22.0\% for MotionGPT3-Qwen versus 90.1\% for ours.
This disparity likely stems from pretraining mismatch: the MotionGPT3 VAE was trained on HumanML3D, which shares AMASS motion-capture sources with BABEL-QA, whereas HuMMan-QA motions were collected via a different pipeline (RGB-D reconstruction).
A feature-level analysis (Appendix~\ref{app:feature_stats}) shows that the resulting input-distribution shift on HuMMan-QA is mild and far smaller than the $-28$ percentage-point accuracy drop, indicating that the gap originates from the VAE's brittleness to cross-source motion data rather than from gross input-quality differences.
This cross-domain fragility is inherent to learned encoders; by contrast, our fully rule-based approach produces consistent representations regardless of the motion data source. Side-by-side qualitative QA examples are provided in Appendix~\ref{app:qualitative_qa}.


\paragraph{Motion captioning.}
On HumanML3D (Table~\ref{tab:main}), our SMD-based approach achieves the best results on all metrics except R@1. Against the strongest prior method MotionGPT3~\citep{chen2024motiongpt3}, we improve R-Precision at every level and MM-Distance, indicating tighter text-motion semantic alignment. The gains in text generation quality are also substantial: $+17\%$ relative on BLEU@4, $+31\%$ on CIDEr, showing higher consensus with multiple references and closer stylistic alignment to human annotations. BERTScore improves from 35.23 to 45.58, confirming stronger semantic similarity at the token embedding level. The same-backbone controlled baseline MotionGPT3-Qwen (Qwen2.5-7B) consistently trails our approach across all metrics, and scaling its projection MLP to 32/64/128 tokens (Appendix~\ref{app:encoder_sweep}) does not close the gap because the larger MLPs overfit on the limited training data. SMD also outperforms MG-MotionLLM~\citep{wu2025mgmotionllm} on the fine-grained motion-to-detailed-text task at both sequence and snippet levels, analyzed in Appendix~\ref{app:finegrained}. Side-by-side qualitative captioning examples are provided in Appendix~\ref{app:qualitative_caption}.

\subsection{Ablation Studies}

\paragraph{Number of joints.}
Table~\ref{tab:joints} varies how many joints are included in SMD affects the performance.``None'' uses only global trajectory, ``Top-$K$'' selects the $K$ joints with the largest angular displacement per motion, and ``All-26'' includes all joints. QA and captioning respond differently. For QA, Top-3 is already best (73.3\%/91.0\%), as focusing on the most active joints filters out noise from static joints. For captioning, retrieval metrics rise monotonically with more joints (R@1: 0.452 with trajectory-only $\to$ 0.584 with All-26), since open-ended generation benefits from richer full-body descriptions. The optimal SMD granularity is thus task-dependent.

\begin{table}[tb]
\centering
\caption{Effect of the number of joints included in SMD. All experiments used Qwen2.5-7B.
}
\label{tab:joints}
\resizebox{\textwidth}{!}{%
\begin{tabular}{lcccccccccccc}
\toprule
Joints & Tokens & BABEL-QA & HuMMan-QA & R@1$\uparrow$ & R@2$\uparrow$ & R@3$\uparrow$ & MMD$\downarrow$ & B@1$\uparrow$ & B@4$\uparrow$ & R-L$\uparrow$ & CIDEr$\uparrow$ & BS$\uparrow$ \\
\midrule
None & $\sim$0.6K & 56.2 & 67.4 & .452 & .626 & .720 & 3.43 & 48.19 & 11.87 & 24.01 & 21.69 & 29.85 \\
Top-3 & $\sim$1K & \textbf{73.3} & \textbf{91.0} & .547 & .754 & .842 & 2.59 & 61.74 & 22.25 & 47.52 & 52.28 & 44.94 \\
Top-5 & $\sim$1.5K & 68.2 & 90.8 & .563 & .757 & .843 & 2.49 & 61.24 & 22.20 & 47.32 & 51.08 & 44.48 \\
Top-10 & $\sim$2.2K & 66.2 & 89.3 & .569 & .773 & .860 & 2.44 & 61.66 & 22.41 & 47.45 & 51.82 & 44.97 \\
Top-20 & $\sim$3.5K & 69.2 & 89.1 & .582 & .793 & .878 & 2.37 & \textbf{63.48} & \textbf{23.07} & \textbf{48.30} & \textbf{56.03} & \textbf{46.22} \\
All-26 & $\sim$4K & 66.7 & 90.1 & \textbf{.584} & \textbf{.794} & \textbf{.883} & \textbf{2.35} & 63.45 & 22.67 & 47.80 & 53.16 & 45.58 \\
\bottomrule
\end{tabular}
}
\end{table}

\begin{table}[t]
\centering
\caption{Effect of trajectory representation in the SMD (All-26 joints, Qwen2.5-7B).}
\label{tab:trajectory}
\resizebox{\textwidth}{!}{%
\begin{tabular}{lccccccccccc}
\toprule
Trajectory & BABEL-QA & HuMMan-QA & R@1$\uparrow$ & R@2$\uparrow$ & R@3$\uparrow$ & MMD$\downarrow$ & B@1$\uparrow$ & B@4$\uparrow$ & R-L$\uparrow$ & CIDEr$\uparrow$ & BS$\uparrow$ \\
\midrule
None & 64.9 & \textbf{90.1} & .574 & .778 & .868 & 2.41 & 62.26 & 22.11 & 47.69 & \textbf{53.34} & \textbf{45.62} \\
Egocentric & \textbf{66.9} & 89.3 & .558 & .762 & .852 & 2.53 & 62.20 & 21.60 & 47.17 & 50.09 & 44.94 \\
Absolute & 66.7 & \textbf{90.1} & \textbf{.584} & \textbf{.794} & \textbf{.883} & \textbf{2.35} & \textbf{63.45} & \textbf{22.67} & \textbf{47.80} & 53.16 & 45.58 \\
\bottomrule
\end{tabular}
}
\end{table}
\paragraph{Trajectory representation.}
Table~\ref{tab:trajectory} compares three ways to encode global trajectory: none, egocentric (body-relative directions), and absolute (world coordinates).
Absolute trajectory achieves the best overall performance, particularly on retrieval metrics (R@1 = 0.584).
Notably, removing trajectory entirely does not cause a dramatic drop: BABEL-QA accuracy decreases by only 1.8 points and captioning CIDEr remains comparable (53.34 vs.\ 53.16).
This may suggest that the model can infer some aspects of global movement from joint angle patterns alone (e.g., cyclic hip and knee flexion implies walking), although explicit trajectory still provides a useful complementary signal.

\paragraph{Sensitivity to SMD conversion parameters.}
Table~\ref{tab:smd_sensitivity} sweeps the three SMD conversion parameters---angle segmentation threshold $\delta$, smoothing window $w$, and trajectory position threshold $\tau_p$. All three are stable: BABEL-QA varies by at most 4 points and captioning R@1 by at most 0.08 across tested settings. A few off-default values (e.g., $\delta = 3$\textdegree{}, $w = 11$) edge out the defaults on R@1/CIDEr, so light per-task tuning may yield modest gains. Overall, SMD is not a brittle hand-tuned prompt but a principled representation that degrades gracefully under parameter perturbation.


\begin{table}[tb]
\centering
\caption{Sensitivity to SMD conversion parameters (All-26 joints, absolute trajectory, Qwen2.5-7B). The table is organized into three sub-blocks, each varying a single parameter while holding the others at their defaults.
}
\label{tab:smd_sensitivity}
\resizebox{\textwidth}{!}{%
\begin{tabular}{lccccccccccc}
\toprule
Setting & BABEL-QA & HuMMan-QA & R@1$\uparrow$ & R@2$\uparrow$ & R@3$\uparrow$ & MMD$\downarrow$ & B@1$\uparrow$ & B@4$\uparrow$ & R-L$\uparrow$ & CIDEr$\uparrow$ & BS$\uparrow$ \\
\midrule
\multicolumn{12}{l}{\textit{Segmentation threshold $\delta$}} \\
3\textdegree{} & 70.2 & 88.7 & \textbf{.596} & \textbf{.796} & .882 & \textbf{2.24} & 62.49 & 22.68 & 47.92 & \textbf{55.20} & 46.02 \\
\textbf{5\textdegree{}} (default) & 66.7 & \textbf{90.1} & .584 & .794 & \textbf{.883} & 2.35 & 63.45 & 22.67 & 47.80 & 53.16 & 45.58 \\
10\textdegree{} & 68.2 & \textbf{90.1} & .544 & .751 & .846 & 2.56 & 62.47 & 22.40 & 47.25 & 50.76 & 44.04 \\
15\textdegree{} & \textbf{70.5} & 88.0 & .580 & .777 & .862 & 2.38 & \textbf{63.50} & \textbf{23.25} & \textbf{48.12} & 53.74 & \textbf{46.03} \\
\midrule
\multicolumn{12}{l}{\textit{Smoothing window $w$ (frames)}} \\
3 & 69.5 & \textbf{90.8} & \textbf{.608} & .802 & .886 & \textbf{2.21} & 63.03 & 22.66 & 47.78 & 54.03 & 45.84 \\
5 & \textbf{71.0} & 89.1 & .591 & .791 & .876 & 2.32 & 62.64 & 22.47 & 47.35 & 52.58 & 45.15 \\
\textbf{7} (default) & 66.7 & 90.1 & .584 & .794 & .883 & 2.35 & \textbf{63.45} & \textbf{22.67} & 47.80 & 53.16 & 45.58 \\
11 & 66.7 & \textbf{90.8} & \textbf{.608} & \textbf{.808} & \textbf{.890} & 2.24 & 62.56 & 22.65 & \textbf{48.04} & \textbf{55.53} & \textbf{46.05} \\
\midrule
\multicolumn{12}{l}{\textit{Trajectory position threshold $\tau_p$ (m)}} \\
0.01 & 69.5 & \textbf{91.7} & .527 & .739 & .833 & 2.54 & 62.76 & 22.40 & 47.70 & 50.69 & 44.46 \\
\textbf{0.03} (default) & 66.7 & 90.1 & \textbf{.584} & \textbf{.794} & \textbf{.883} & \textbf{2.35} & \textbf{63.45} & \textbf{22.67} & 47.80 & 53.16 & \textbf{45.58} \\
0.05 & \textbf{70.2} & 90.8 & .543 & .749 & .831 & 2.55 & 62.34 & 22.27 & 47.39 & 50.19 & 44.08 \\
0.10 & 68.7 & 90.8 & .572 & .781 & .871 & 2.39 & 63.17 & 22.64 & \textbf{48.19} & \textbf{54.04} & 45.79 \\
\bottomrule
\end{tabular}
}
\end{table}
\begin{table}[tb]
\centering
\caption{
Performance with and without LoRA fine-tuning on Qwen2.5-7B using Top-3 SMD.}
\label{tab:zeroshot}
\resizebox{\textwidth}{!}{%
\begin{tabular}{lccccccccccc}
\toprule
Setting & BABEL-QA & HuMMan-QA & R@1$\uparrow$ & R@2$\uparrow$ & R@3$\uparrow$ & MMD$\downarrow$ & B@1$\uparrow$ & B@4$\uparrow$ & R-L$\uparrow$ & CIDEr$\uparrow$ & BS$\uparrow$ \\
\midrule
Random guess & $\sim$11.6 & $\sim$10.0 & $\sim$.031 & $\sim$.063 & $\sim$.094 & --- & --- & --- & --- & --- & --- \\
Zero-shot & 35.6 & 31.7 & .075 & .147 & .221 & 6.86 & 19.76 & 1.14 & 18.45 & 0.86 & 8.26 \\
Fine-tuned & \textbf{73.3} & \textbf{91.0} & \textbf{.547} & \textbf{.754} & \textbf{.842} & \textbf{2.59} & \textbf{61.74} & \textbf{22.25} & \textbf{47.52} & \textbf{52.28} & \textbf{44.94} \\
\bottomrule
\end{tabular}
}
\end{table}

\paragraph{Zero-shot performance.}
Table~\ref{tab:zeroshot} compares performance with and without LoRA fine-tuning.
Without fine-tuning (zero-shot), the LLM achieves 35.6\% on BABEL-QA and 31.7\% on HuMMan-QA using Top-3 SMD, substantially above random chance (10\% for 10 options); answers are extracted with a lightweight regex that strips conversational prefixes (e.g., ``The answer is'', ``Option:'') before exact-string matching, so the score reflects model reasoning rather than formatting failures. This indicates that the LLM can partially interpret biomechanical descriptions in SMD even without task-specific training.
For captioning, the zero-shot model generates SMD-grounded but overly verbose descriptions, correctly identifying low-level movement components (e.g., lateral sway, torso rotation, arm swinging) but failing to summarize them into a concise action label---see Appendix~\ref{app:zeroshot_caption} for walking and waltz examples. This suggests that while the LLM can extract low-level motion cues from SMD without training, LoRA fine-tuning is needed to map biomechanical patterns to action-level semantics and to learn a concise captioning style. What SMD eliminates is the need for a learned motion encoder and multi-stage alignment pipeline, not the need for any task adaptation.

\subsection{Backbone Portability}
\label{sec:backbone}

Because SMD is plain text, switching the LLM backbone only requires retraining a lightweight LoRA adapter ($\sim$40M parameters)---no motion encoder, projection layer are needed. To demonstrate portability, we train 8 LLMs across 6 model families (3B--14B parameters) using the Top-3 SMD variant ($\sim$1{,}000 tokens/motion). Table~\ref{tab:backbone} shows consistent performance across architectures: all models achieve $>$63\% on BABEL-QA, $>$82\% on HuMMan-QA, R@1 in 0.517--0.563, and CIDEr in 49.23--54.33. Within the Qwen2.5 family, scale helps (3B $\to$ 7B $\to$ 14B), and newer generations (Qwen3-8B, Qwen3.5-9B) match or surpass Qwen2.5-14B, consistent with more capable LLMs extracting more from the same SMD input; even the smallest model (Gemma3-4B) remains competitive.

While switching backbone is convenient, the trade-off is a longer input sequence ($\sim$1{,}000--4{,}000 SMD tokens vs.\ $\sim$256 motion tokens for VAE/VQ-VAE methods), yielding roughly 1 sample/s inference on a single H200; see Appendix~\ref{app:cost} for the full cost breakdown.

\begin{table}[tb]
\centering
\caption{Backbone portability across 8 LLMs from 6 model families. All models use the same Top-3 SMD input ($\sim$1{,}000 tokens) and identical LoRA configuration.}
\label{tab:backbone}
\resizebox{\textwidth}{!}{%
\begin{tabular}{lcccccccccc}
\toprule
LLM & Params & BABEL-QA & HuMMan-QA & R@1$\uparrow$ & MMD$\downarrow$ & B@1$\uparrow$ & B@4$\uparrow$ & R-L$\uparrow$ & CIDEr$\uparrow$ & BS$\uparrow$ \\
\midrule
Qwen2.5-3B & 3B & 64.6 & 89.3 & .517 & 2.72 & 60.79 & 20.80 & 47.23 & 50.63 & 44.18 \\
Qwen2.5-7B & 7B & \textbf{73.3} & 91.0 & .547 & 2.59 & 61.74 & 22.25 & 47.52 & 52.28 & 44.94 \\
Qwen2.5-14B & 14B & 71.5 & 88.7 & .540 & 2.61 & 61.83 & 21.85 & 47.35 & 49.23 & 44.55 \\
Qwen3-8B & 8B & 69.7 & 89.3 & .541 & 2.49 & 60.75 & \textbf{23.30} & \textbf{47.84} & \textbf{54.33} & \textbf{45.60} \\
Qwen3.5-9B & 9B & 72.8 & \textbf{91.7} & .541 & 2.61 & 60.43 & 22.43 & 47.56 & 51.39 & 45.17 \\
Gemma3-4B & 4B & 66.2 & 88.4 & .552 & 2.55 & \textbf{61.77} & 22.85 & 47.64 & 52.26 & 45.28 \\
LLaMA-3.1-8B & 8B & 63.1 & 89.1 & .535 & 2.62 & 59.57 & 22.02 & 46.85 & 49.89 & 44.34 \\
GLM-4-9B & 9B & 67.9 & 82.9 & \textbf{.563} & \textbf{2.44} & 60.86 & 22.75 & 47.58 & 53.20 & 45.10 \\
\bottomrule
\end{tabular}
}
\end{table}


\subsection{Interpretability}
\label{sec:interpretability}

\begin{figure}[tb]
\centering
\includegraphics[width=0.85\textwidth]{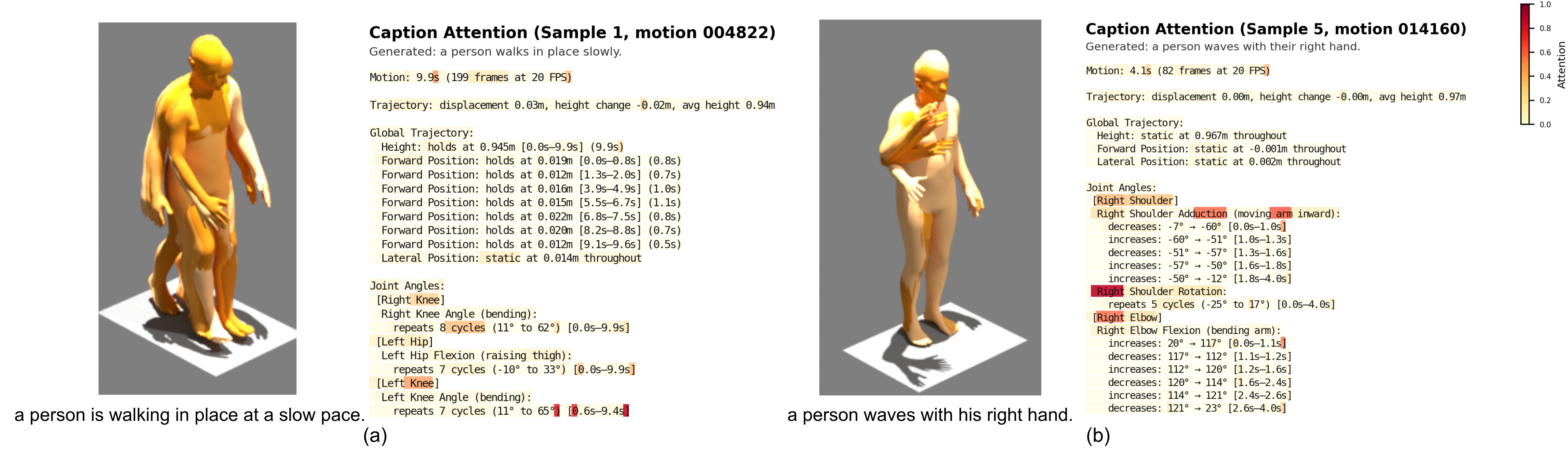}
\caption{Attention heatmaps for two captioning examples. Background color on the SMD text indicates attention weight (yellow = low, red = high). (a) For ``walking in place'' the model attends to trajectory segments and cyclic joint angle patterns. (b) For ``waving with right hand'' the model focuses on the Right Shoulder and Right Elbow segments.}
\label{fig:attention_caption}
\end{figure}

A distinctive advantage of SMD is its human-readable, text-based motion representation, which enables straightforward attention-based interpretability---something far harder with learned representations that compress motion into a few opaque latent tokens.
We aggregate the attention weights over the SMD input across all layers of fine-tuned Qwen2.5-7B. Figure~\ref{fig:attention_caption} shows heatmaps for two captioning examples.
In both examples, the model selectively focuses on the SMD sections that are most relevant to the generated caption, demonstrating interpretable reasoning traces.

\section{Conclusion}
\label{sec:conclusion}

We have presented SMD, a rule-based conversion from human motion to structured text that enables LLMs to perform motion understanding without learned encoders or cross-modal alignment. With only lightweight LoRA fine-tuning, SMD surpasses prior state-of-the-art on both motion QA and captioning, transfers consistently across 8 LLMs from 6 families at 2–8 GPU-hours per adapter, and offers built-in interpretability via attention over human-readable tokens.

\paragraph{Limitations and Future Work.}
The primary limitation is inference latency: the $\sim$4{,}000 tokens of All-26 SMD are roughly $15\times$ longer than the $\sim$256 motion tokens used by VAE-based methods, raising per-sample inference time (training cost is unaffected, being a single LoRA run). The rule-based conversion also targets a fixed set of 26 biomechanical angles over 22 SMPL joints, which may miss motion nuances in more detailed skeleton formats (e.g., hand and finger articulation). Promising future directions include learned alternatives that generate SMD-style text end-to-end from motion sequences, motion-adaptive segmentation thresholds to better capture both very slow and highly explosive motions that our current fixed thresholds may smooth over, and extending SMD beyond understanding to motion generation and editing.

{
\small
\bibliographystyle{plainnat}
\bibliography{references}
}

\appendix

\section{Method Details}

\subsection{Complete list of joint angles}
\label{app:angles}
Our joint angle computation follows the biomechanical framework of \citet{zhang2026jami, schlegel2024joint, wu2002isb, wu2005isb}, which defines angles by projecting bone vectors onto anatomical reference planes.

At each time frame, we establish a body-local coordinate frame using three reference points---the pelvis (origin), the left hip, and the right hip---following the convention $\mathbf{e}_x = \text{forward}$, $\mathbf{e}_y = \text{up}$, $\mathbf{e}_z = \text{lateral}$ (right-pointing): the lateral axis $\mathbf{e}_z$ is the unit vector from the left hip to the right hip; the vertical axis $\mathbf{e}_y$ is the unit vector from the hip midpoint to the pelvis, orthogonalized against $\mathbf{e}_z$; and the forward axis $\mathbf{e}_x = \mathbf{e}_y \times \mathbf{e}_z$ completes the right-handed system, with its sign chosen so that it points anteriorly. This body-local frame defines the subject's overall facing direction and serves as the root of the kinematic chain.

Individual joint angles are then computed within a hierarchy of joint-local coordinate frames, each defined relative to its parent joint along the kinematic chain. For example, hip angles are expressed in the pelvis frame; knee flexion in the hip (femur) frame; ankle angles in the knee (tibia) frame; shoulder angles in the lumbar-spine frame; and elbow flexion in the shoulder (upper-arm) frame. Expressing each angle in its parent frame follows standard ISB conventions~\citep{wu2002isb, wu2005isb} and isolates the motion of each joint from the orientation of upstream segments, so that, for instance, knee flexion measures only the bending of the knee regardless of how the hip or torso is oriented. The body-local root frame is derived from intrinsic body landmarks, and every joint-local frame is expressed relative to its parent along the kinematic chain. As a result, all joint angles encode only relative segment orientations and are invariant to the subject's global position and orientation.

\begin{figure}[htbp]
\centering
\includegraphics[width=0.4\textwidth]{figs/fig_angles.pdf}
\caption{Illustration of the body-local and joint-local coordinate frames used for joint angle computation. The body-local frame (attached to the pelvis) is first established from the pelvis, left hip, and right hip, and defines the subject's overall facing direction. Joint-local frames are then attached to each of the 13 body part groups, organized into a kinematic chain so that each joint angle is expressed relative to its parent segment (e.g., knee flexion in the hip frame, elbow flexion in the shoulder frame). The three colored axes at each joint indicate forward (red), vertical (green), and lateral (blue). Dashed lines at the knees, elbows, and ankles show the initial (zero-flexion) positions against which flexion angles are measured, and the magenta arrow illustrates the rotation convention for joints with a twist degree of freedom (e.g., shoulder, hip). The global coordinate system (bottom-right) is shown for reference.}
\label{fig:angles_illustration}
\end{figure}

The 26 angles are organized into 13 body part groups, listed below along with their parent reference frame (labeled in Figure~\ref{fig:angles_illustration}):
\begin{enumerate}[leftmargin=*,topsep=2pt,itemsep=1pt]
\item \textbf{Pelvis} (3 angles, in the global frame): Tilt (forward/backward lean, projected onto the sagittal plane), List (lateral tilt, projected onto the coronal plane), and Rotation (yaw of the body forward direction).
\item \textbf{Lumbar Spine} (3 angles, in the pelvis frame): Extension, Lateral Bending, and Rotation, computed from the spine1-to-spine3 vector.
\item \textbf{Neck} (2 angles, in the lumbar-spine frame): Flexion (nodding) and Lateral Tilt, computed from the neck-to-head vector.
\item \textbf{Hip L/R} (3 angles each, in the pelvis frame): Flexion (sagittal plane), Adduction (coronal plane), and Rotation (femur twist).
\item \textbf{Knee L/R} (1 angle each, in the hip frame): Flexion, the angle between the femur and tibia vectors.
\item \textbf{Ankle L/R} (1 angle each, in the knee frame): Dorsi/plantarflexion, the angle between the tibia and foot vectors.
\item \textbf{Shoulder L/R} (3 angles each, in the lumbar-spine frame): Flexion (sagittal plane), Adduction (coronal plane), and Rotation (upper-arm twist).
\item \textbf{Elbow L/R} (1 angle each, in the shoulder frame): Flexion, the angle between the upper-arm and forearm vectors.
\end{enumerate}

All angles are reported in degrees.
Sign conventions follow standard biomechanical practice: flexion is positive, extension is negative; adduction is positive, abduction is negative.
Wrappable angles (pelvis rotation, hip rotation, shoulder rotation) are unwrapped to avoid discontinuities at $\pm 180$\textdegree{} before temporal segmentation. The pipeline assumes complete 3D joint positions from SMPL-format input; missing or noisy joints would need to be imputed upstream of SMD, and 2D inputs would require an off-the-shelf 3D pose lifter.

\paragraph{Axial twist via a secondary reference vector.}
A single 3D bone vector has only two directional degrees of freedom, so an axial twist around its longitudinal axis cannot be computed from the bone alone---a secondary reference vector pointing out of the bone axis is required. We construct a parent-segment-aligned local frame using the joint's flexion and adduction angles, then measure the angle of a secondary vector around the segment axis: hip rotation uses the tibia (knee-to-ankle) vector inside the femur-aligned frame; lumbar rotation uses the neck-to-right-shoulder vector inside the spine-aligned frame; and shoulder rotation uses the forearm (elbow-to-wrist) vector inside the upper-arm-aligned frame. In each case, the twist angle is the signed angle of the projection of the secondary vector onto the plane orthogonal to the segment's longitudinal axis.

\paragraph{Segmentation and cyclic detection.}
After calculated all 26 joint angles, each joint angle (and trajectory-axis) time series is first smoothed with a moving-average filter of window $w$, then segmented by peak/valley detection with a minimum amplitude $\delta$, yielding an alternating sequence of ``increases'' and ``decreases'' intervals.

Table~\ref{tab:smd_vocab} lists the descriptive phrase that appears verbatim in the SMD text for each global trajectory axis and joint angle. The anatomical paraphrase in parentheses (e.g., ``raising thigh'' for hip flexion, ``bending arm'' for elbow flexion) makes each quantity immediately interpretable to an LLM relying on its pretrained world knowledge of human body mechanics, without any learned alignment.

\begin{table}[htbp]
\centering
\small
\caption{Descriptive phrases used in the SMD for each global trajectory axis and joint angle. The Global Trajectory block lists the verbs used for positive/negative/static segments along each axis. The Joint Angles block lists the display phrase that appears verbatim in the SMD text; each segment of a joint angle is additionally tagged with one of ``increases'', ``decreases'', ``holds at'', or ``repeats $N$ cycles'' (e.g., ``Left Hip Flexion (raising thigh): increases $3\text{\textdegree} \to 81\text{\textdegree}$ [0.0s--0.9s]''). The 13 body-part groups (shown as bracketed headers) match the section headers that appear in the assembled SMD string.}
\label{tab:smd_vocab}
\begin{tabular}{l l}
\toprule
\textbf{Global trajectory axis} & \textbf{Direction verbs (positive / negative / static)} \\
\midrule
Height & rises / lowers / holds at \\
Forward Position & moves forward / moves backward / holds at \\
Lateral Position & moves left / moves right / holds at \\
Body Rotation & turns left / turns right / holds at \\
\midrule
\textbf{Body-part group (13)} & \textbf{Joint-angle display phrase (26)} \\
\midrule
\multirow{3}{*}{[Pelvis]}
 & Pelvis Tilt (forward/backward lean) \\
 & Pelvis List (lateral tilt) \\
 & Pelvis Rotation (turning) \\
\midrule
\multirow{3}{*}{[Lumbar Spine]}
 & Lumbar Extension (bending forward/backward) \\
 & Lumbar Lateral Bending (side bend) \\
 & Lumbar Rotation (twisting) \\
\midrule
\multirow{2}{*}{[Neck]}
 & Neck Flexion (nodding) \\
 & Neck Lateral Tilt (tilting head sideways) \\
\midrule
\multirow{3}{*}{[Right Hip]}
 & Right Hip Flexion (raising thigh) \\
 & Right Hip Adduction (moving leg inward) \\
 & Right Hip Rotation \\
\midrule
{[Right Knee]} & Right Knee Angle (bending) \\
{[Right Ankle]} & Right Ankle Angle (dorsi/plantarflexion) \\
\midrule
\multirow{3}{*}{[Left Hip]}
 & Left Hip Flexion (raising thigh) \\
 & Left Hip Adduction (moving leg inward) \\
 & Left Hip Rotation \\
\midrule
{[Left Knee]} & Left Knee Angle (bending) \\
{[Left Ankle]} & Left Ankle Angle (dorsi/plantarflexion) \\
\midrule
\multirow{3}{*}{[Right Shoulder]}
 & Right Shoulder Flexion (raising arm forward) \\
 & Right Shoulder Adduction (moving arm inward) \\
 & Right Shoulder Rotation \\
\midrule
{[Right Elbow]} & Right Elbow Flexion (bending arm) \\
\midrule
\multirow{3}{*}{[Left Shoulder]}
 & Left Shoulder Flexion (raising arm forward) \\
 & Left Shoulder Adduction (moving arm inward) \\
 & Left Shoulder Rotation \\
\midrule
{[Left Elbow]} & Left Elbow Flexion (bending arm) \\
\bottomrule
\end{tabular}
\end{table}

\begin{figure}[htbp]
\centering
\includegraphics[width=0.7\textwidth]{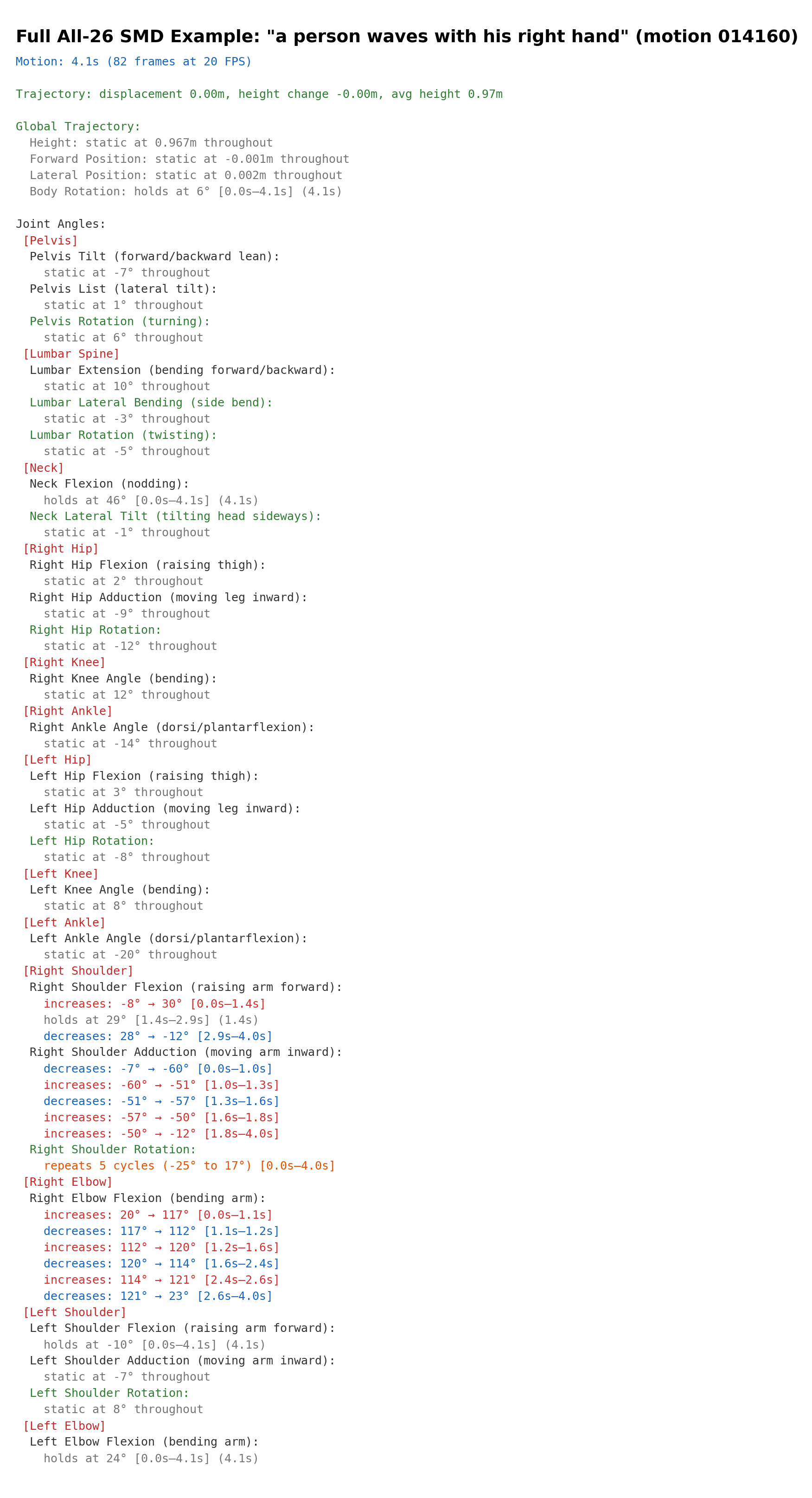}
\caption{Complete All-26 SMD for motion 014160 (``a person waves with his right hand'' 4.1s). Color coding: body part headers (dark red), increases (red), decreases (blue), static/holds (gray), repeats (orange), trajectory (green). The Right Shoulder and Right Elbow sections show active movement while all other joints remain static.}
\label{fig:smd_all26}
\end{figure}

A complete All-26 SMD example for the motion ``a person waves with his right hand'' is shown in Figure~\ref{fig:smd_all26}.
The full text contains 87 lines covering all 13 body part groups: most joints are static (gray), while the Right Shoulder and Right Elbow show active movement patterns (red/blue/orange) corresponding to the waving action. For the Top-$K$ variant of SMD, joints are ranked by the peak-to-trough range $\max_t \theta_k^{(t)} - \min_t \theta_k^{(t)}$ over the motion, and the $K$ joints with the largest ranges are kept.

\subsection{Prompt structure}
\label{app:prompt_format}

\begin{figure}[h]
\centering
\includegraphics[width=\textwidth]{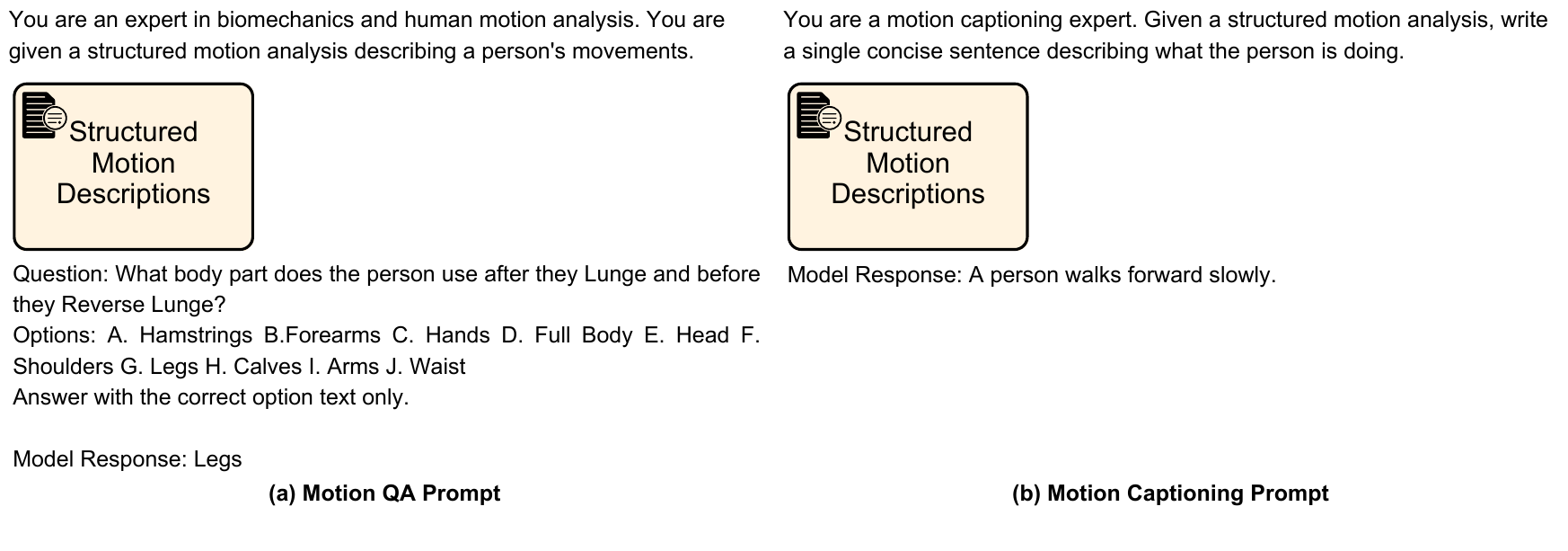}
\caption{Prompt structure for (a) motion QA and (b) motion captioning.}
\label{fig:prompt_format}
\end{figure}

Figure~\ref{fig:prompt_format} shows the full prompt template used during LoRA fine-tuning and inference. For motion QA (a), the prompt consists of a system instruction describing the task role, the SMD $\mathcal{S}$, the question text $q$, and the candidate answer options; the target output is the text of the correct option. For motion captioning (b), the prompt consists of a system instruction and the SMD $\mathcal{S}$; the target output is a natural-language caption describing the motion (e.g., ``a person walks forward slowly''). During training, only the target response tokens contribute to the loss (all prompt tokens are masked).

\section{Additional Analyses}
\subsection{Open-vocabulary QA on BABEL-QA}
\label{app:openvocab}

The standardized 10-option format used in Table~\ref{tab:main} bounds the answer to a discrete candidate set, which is convenient for comparing different methods. However, since MotionLLM~\citep{chen2024motionllm} does not provide checkpoints or training data, we were unable to retrain and evaluate MotionLLM~\citep{chen2024motionllm} using the 10-option versions of BABEL-QA and HuMMan-QA. Therefore, the comparison with MotionLLM in the table~\ref{tab:main} is not fair. To enable a direct comparison with MotionLLM~\citep{chen2024motionllm}, which reports open-vocabulary numbers in its original paper, we additionally evaluate our SMD-based approach on BABEL-QA and HuMMan-QA in a free-form generation setting: the prompt drops the candidate options entirely, and the model must generate the answer phrase autoregressively. We measure exact-match accuracy after a lightweight regex pass that strips conversational prefixes (e.g., ``The answer is'', ``Option:''), lowercases the result, and removes leading/trailing punctuation, matching the protocol used for our zero-shot evaluation in Section~\ref{sec:exp}. We use the same Qwen2.5-7B + LoRA checkpoint as in Table~\ref{tab:backbone} without any re-fine-tuning and Top-3 joints SMD as input. This open-vocabulary evaluation is fair to all baselines: MotionLLM, IMoRe, and NSPose all train with the candidate options visible to the model, so removing the options at test time is a strictly harder regime that the comparison points were never optimized for.

Removing the candidate options at test time costs our model only 8.2 and 7.1 percentage points on BABEL-QA and HuMMan-QA respectively, and our open-vocabulary accuracy on BABEL-QA (65.1\%) still exceeds the original MotionLLM number (43.6\%) by more than 20 points. This indicates that LoRA-fine-tuned SMD does not rely on having the candidate options visible.

\begin{table}[h]
\centering
\caption{Open-vocabulary motion QA accuracy (\%). $^\dagger$Number from the MotionLLM paper, which reports the original open-vocabulary protocol.}
\label{tab:openvocab}
\resizebox{0.85\textwidth}{!}{%
\begin{tabular}{lccc}
\toprule
Method & Backbone & BABEL-QA & HuMMan-QA \\
\midrule
MotionLLM$^\dagger$~\citep{chen2024motionllm} (open-vocab) & Vicuna-13B & 43.6 & --- \\
Ours (10-option, for reference) & Qwen2.5-7B & 73.3 & 91.0 \\
Ours (open-vocab) & Qwen2.5-7B & 65.1 & 83.9 \\
\bottomrule
\end{tabular}
}
\end{table}


\subsection{Zero-shot captioning examples}
\label{app:zeroshot_caption}

\begin{figure}[htbp]
\centering
\includegraphics[width=\textwidth]{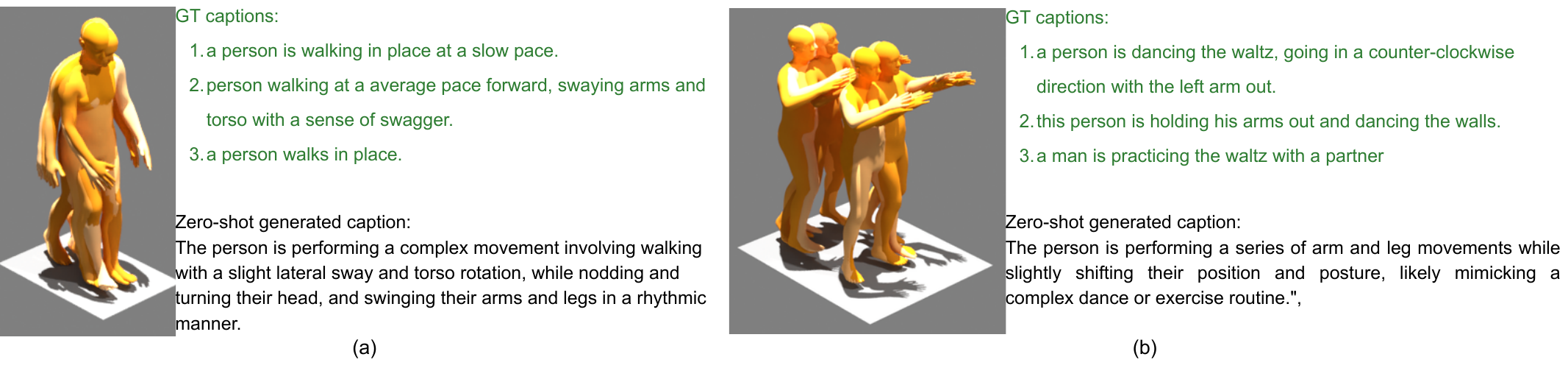}
\caption{Zero-shot captioning examples. (a) Walking: the model identifies movement components (lateral sway, torso rotation, arm swinging) but produces a verbose description rather than the concise ``a person walks in place.'' (b) Waltz: the model recognizes arm and leg movements but describes the action generically as ``a complex dance or exercise routine'' failing to identify the waltz.}
\label{fig:zeroshot_caption}
\end{figure}

Figure~\ref{fig:zeroshot_caption} illustrates the qualitative behavior of the zero-shot model (pre-fine-tuning) discussed in Section~\ref{sec:exp}. For a walking motion (a), the zero-shot model correctly identifies the movement components (lateral sway, torso rotation, arm swinging) but produces a verbose description rather than the concise ``a person walks in place.'' For a waltz (b), the model recognizes arm and leg movements but fails to identify the high-level action, describing it generically as ``a complex dance or exercise routine'' rather than a waltz. These examples show that the LLM can partially interpret SMD without training, but LoRA fine-tuning is needed to learn the concise captioning style and to map biomechanical patterns to action-level semantics.

\subsection{Feature distribution comparison: BABEL-QA vs HuMMan-QA}
\label{app:feature_stats}

This appendix examines whether the MotionGPT3-Qwen accuracy gap between BABEL-QA ($50.1\%$) and HuMMan-QA ($22.0\%$) can be attributed to a distribution shift in the inputs that the frozen MotionGPT3 VAE receives.

\paragraph{Setup.}
For every BABEL-QA motion (1{,}109 motions, 382{,}647 frames), every HuMMan-QA motion (925 motions, 119{,}501 frames), and a 2{,}000-motion subset of HumanML3D (281{,}940 frames) we compute the 263-dim HumanML3D feature~\citep{guo2022t2m} using the standard MotionGPT3 pipeline, then z-normalize each dimension $d$ with the HumanML3D Mean/Std used for VAE pretraining: $z^{(t)}_d = (f^{(t)}_d - \mu^\text{HML3D}_d) / \sigma^\text{HML3D}_d$. This is exactly the input the VAE receives at inference time. We split the 263 dims into five semantic groups --- root (4 dims), joint\_pos (63 dims), joint\_rot6d (126 dims), joint\_vel (66 dims), foot\_contact (4 dims) --- and report all statistics as group averages. The HumanML3D row in Table~\ref{tab:feature_stats} serves as a sanity check: $|\text{mean}_z| \approx 0$ and $\text{std}_z \approx 1$ confirm that the same pipeline applied to in-domain data correctly reproduces the VAE's training distribution.

\paragraph{Metrics.}
For each dataset we report the per-dim absolute mean $|\text{mean}_z|$ (systematic shift from HumanML3D; $0$ is ideal) and standard deviation $\text{std}_z$ (spread; $1$ matches HumanML3D). To measure how different each QA dataset is from the VAE's training distribution, we report the 1D Wasserstein-1 distance $W_1$ between its per-dim empirical CDFs and HumanML3D's (200-bin histograms over $z \in [-20, 20]$, averaged within a group). $W_1$ is the average distance, in z-score units, required to move probability mass to turn one distribution into the other: $W_1 < 0.5$ is near-identical, $W_1 < 1$ is comparable with mild shift, and $W_1 > 2$ is meaningfully different.

\paragraph{Results.}
Table~\ref{tab:feature_stats} reports per-dataset z-statistics for the three datasets and pairwise Wasserstein-1 distances against HumanML3D. BABEL-QA features are essentially in-distribution for the VAE: the overall $|\text{mean}_z|$ is $0.10$ and the overall $\text{std}_z$ is $0.98$, with a pairwise $W_1 = 0.14$ to HumanML3D --- consistent with BABEL motions being drawn from the same AMASS source. HuMMan-QA exhibits a mild shift: $|\text{mean}_z| = 0.25$, $\text{std}_z = 0.68$, and $W_1 = 0.36$ to HumanML3D, well below the $W_1 > 2$ ``meaningfully different'' threshold but larger than BABEL-QA. This confirms that the RGB-D--reconstructed HuMMan motions occupy a slightly different region of the VAE's input space than the AMASS-derived training distribution, but only modestly so.

\begin{table}[h]
\centering
\small
\caption{Per-dataset z-statistics on the 263-dim HumanML3D features after z-normalization, plus the pairwise Wasserstein-1 distance to HumanML3D. All quantities are averaged within each feature group. Values closer to $|\text{mean}_z|=0$, $\text{std}_z=1$, and $W_1=0$ indicate closer agreement with the VAE pretraining distribution.}
\label{tab:feature_stats}
\resizebox{\linewidth}{!}{
\begin{tabular}{llcccccc}
\toprule
Metric & Dataset & root & joint\_pos & joint\_rot6d & joint\_vel & foot\_contact & overall \\
\midrule
\multirow{3}{*}{$|\text{mean}_z|$}
  & BABEL-QA  & 0.10 & 0.11 & 0.14 & 0.03 & 0.05 & 0.10 \\
  & HuMMan-QA & 0.18 & 0.25 & 0.31 & 0.15 & 0.23 & 0.25 \\
  & HumanML3D & 0.04 & 0.04 & 0.01 & 0.01 & 0.01 & 0.02 \\
\midrule
\multirow{3}{*}{$\text{std}_z$}
  & BABEL-QA  & 0.81 & 1.03 & 1.09 & 0.73 & 0.94 & 0.98 \\
  & HuMMan-QA & 0.47 & 0.94 & 0.65 & 0.49 & 0.69 & 0.68 \\
  & HumanML3D & 1.00 & 0.99 & 0.99 & 1.01 & 1.00 & 1.00 \\
\midrule
\multirow{2}{*}{$W_1$ vs HumanML3D}
  & BABEL-QA  & 0.12 & 0.13 & 0.15 & 0.13 & 0.06 & 0.14 \\
  & HuMMan-QA & 0.35 & 0.35 & 0.41 & 0.28 & 0.24 & 0.36 \\
\bottomrule
\end{tabular}
}
\end{table}

Within the standard 263-dim representation, HuMMan-QA differs from HumanML3D only by acquisition source---HuMMan was collected via RGB-D reconstruction while HumanML3D and BABEL-QA both come from the AMASS MoCap pipeline that the VAE was trained on. The resulting input-distribution shift is modest ($W_1 = 0.36$, far below the $W_1 > 2$ ``meaningfully different'' threshold; for reference, the AMASS-derived BABEL-QA gives $W_1 = 0.14$). Yet MotionGPT3-Qwen's accuracy collapses by 28 percentage points on HuMMan-QA ($50.1\% \to 22.0\%$). The disproportion between this small input shift and the large performance drop exposes a generalization gap inherent to encoder-based methods: a VAE pretrained on a single-source corpus produces unreliable latents on data of the same type but a slightly different source. SMD sidesteps this issue: a rule-based deterministic pipeline produces statistically equivalent text on both datasets, and the LLM (pretrained on a far broader text corpus) is unaffected by the source change.

\section{Use Cases}

\subsection{Cross-skeleton generalization (NTU RGB+D 60)}
\label{app:cross_skeleton}

The 22-joint SMPL skeleton used in Section~\ref{sec:method} is just one of several common motion representations. To probe whether SMD generalizes beyond this topology, we apply it to NTU RGB+D~60~\citep{shahroudy2016ntu}, a Kinect-v2 dataset with 25 joints per frame and 60 action classes. NTU is a strong test of generalization for two reasons: it uses a different sensor (Kinect depth, not optical MoCap) and a different skeleton topology (25 vs.\ 22 joints, with NTU including hand-finger joints absent in our SMPL subset). We use the X-Sub split, which separates training and test subjects.

\paragraph{Joint and frame mapping.}
NTU's 25-joint skeleton already shares a hierarchical structure with SMPL: 19 of the 25 NTU joints have a direct SMPL counterpart (knees, ankles, shoulders, elbows, wrists, hips, neck, head, etc.). Three additional SMPL joints absent from NTU---spine1, left/right collar---are interpolated as midpoints of nearby NTU joints (SpineBase--SpineMid for spine1, Neck--Shoulder for the collars). The four NTU finger and thumb joints are dropped. We additionally flip the lateral axis sign to convert NTU's right-handed camera frame ($x$ pointing to the subject's right) into the HumanML3D convention used by SMD, and downsample from 30~FPS to 20~FPS to match SMD's training-time frame rate.

A subtle but important detail concerns the SMPL pelvis joint. In SMPL/HumanML3D the pelvis sits roughly $9$~cm above the hip midpoint along the spine, so the up-direction $\mathbf{e}_y \propto \text{pelvis} - \text{hip\_mid}$ used in $f_\text{SMD}$'s body-local frame (Section~\ref{sec:smd}, Appendix~\ref{app:angles}) is a clean, well-conditioned vector. NTU's SpineBase joint, by contrast, sits essentially at the hip midpoint ($\approx 1$~mm offset on a typical motion), so mapping SpineBase directly to the SMPL pelvis would make this difference vector dominated by Kinect noise and corrupt every downstream joint angle. We therefore map the SMPL pelvis to NTU's SpineMid (joint~1, $\approx 26$~cm above the hips), which restores the same well-conditioned up direction and produces angle distributions visually consistent with HumanML3D-derived motion. No other modification to $f_\text{SMD}$ is required.

\paragraph{Evaluation protocol.}
We frame action recognition as a 10-option QA: ``What action is the person performing? Options: A.\,\ldots\, J.''---9 distractor classes are sampled from the remaining 59 NTU classes per query (fixed seed) plus the correct one. We compare two regimes:
(i) \textbf{Zero-shot}: we apply our All-26 SMD checkpoint trained on BABEL-QA + HuMMan-QA (Table~\ref{tab:main}), with no NTU-specific training.
(ii) \textbf{Fine-tuned (1~epoch)}: we generate SMD All-26 for the 38{,}165 training motions of NTU60 X-Sub and continue LoRA fine-tuning the same Qwen2.5-7B base for one epoch, using the same hyperparameters as Section~\ref{sec:exp}.

\begin{table}[h]
\centering
\caption{Action recognition on NTU RGB+D~60 (X-Sub split, 10-option format). Random-guess accuracy is $10\%$. The zero-shot row uses our BABEL/HuMMan-trained checkpoint with no NTU exposure; the fine-tuned row uses the same backbone with one epoch of LoRA fine-tuning on the NTU train split.}
\label{tab:ntu_zeroshot}
\resizebox{0.85\textwidth}{!}{%
\begin{tabular}{lcc}
\toprule
Setting & Training data & Top-1 (\%) \\
\midrule
Random guess & --- & 10.0 \\
Ours (SMD All-26, zero-shot)               & BABEL-QA + HuMMan-QA        & 20.5 \\
\textbf{Ours (SMD All-26, 1-epoch FT)}     & NTU60 training set & \textbf{83.8} \\
\bottomrule
\end{tabular}
}
\end{table}

\paragraph{Discussion.}
First, the zero-shot number ($20.5\%$, $\sim 2\times$ chance) is modest compared with our in-distribution numbers in Table~\ref{tab:main}. A per-class breakdown explains why: the model performs strongly on broad locomotor and gestural actions that resemble its BABEL/HuMMan training distribution---e.g., \textit{throw} and \textit{hopping} reach $100\%$, \textit{jump up}, \textit{kicking something}, \textit{staggering}, \textit{falling}, and \textit{walking apart from each other} reach $80\%$---but collapses to $0\%$ on two failure groups: (a)~everyday activities never seen in pre-training, such as \textit{eat meal}, \textit{brushing teeth}, \textit{wear jacket}, \textit{put on glasses}, \textit{check time from watch}, \textit{typing on keyboard}, and \textit{reading}; and (b)~activities that are intrinsically hard to disambiguate from skeleton alone because they hinge on object interaction or fine-grained finger movement, such as \textit{tear up paper}, \textit{wipe face}, \textit{touch head/neck/back}, and \textit{pat on back of other person}. The first group reflects a real limitation of our pre-training corpus: training only on BABEL-QA and HuMMan-QA exposes the LLM to a relatively narrow action vocabulary (locomotion + body-part gestures + fitness), and large-scale multi-task fine-tuning across more diverse activity datasets is a natural avenue for future work.

Second, after just one epoch of LoRA fine-tuning on the NTU60 X-Sub training split, the same model reaches $83.8\%$ top-1 on the held-out test split, a $63$-point jump over zero-shot. The SMD conversion pipeline, prompt template, and architecture are all unchanged from Section~\ref{sec:exp}; only the LoRA adapter weights are updated. This confirms two things: (i)~the SMD biomechanical text representation transfers to a different skeleton format and acquisition pipeline once the LLM is exposed to the new action vocabulary, and (ii)~SMD's conversion stack is data-efficient, since a single epoch on a non-AMASS, non-MoCap dataset is enough to recover most of the gap on previously unseen everyday actions.

\subsection{Fine-grained motion captioning}
\label{app:finegrained}

MG-MotionLLM~\citep{wu2025mgmotionllm} introduces a fine-grained motion-to-detailed-text (m2dt) task in which the caption must describe per-body-part movements over time rather than only the overall action label. SMD is a natural fit for this task: its per-joint, per-segment text already carries the fine-grained structure that the task requires, so the LLM can directly transcribe and combine joint-level information rather than having to recover it from a compressed motion latent.

\paragraph{Dataset.}
We evaluate on the FineMotion test set~\citep{wu2025mgmotionllm}, which re-annotates each HumanML3D motion with detailed body-movement descriptions over time. Each motion sequence is divided into fixed-length snippets of 0.5\,s, and each snippet is paired with an automatically generated, fine-grained description of the movements of individual body parts during that interval (e.g., ``\textit{0.0s--0.5s: Move your right hand to your right shoulder. 0.5s--1.0s: Move your right elbow down to your chest. Extend your left arm out to the side slightly.}''). Snippets are joined into a single ground-truth document for the whole motion via a special $\langle \texttt{SEP} \rangle$ token; empty snippets are replaced by $\langle \texttt{Motionless} \rangle$. In total, FineMotion contains 420{,}968 snippets across the HumanML3D motion set.

\paragraph{Two-level evaluation protocol.}
Following MG-MotionLLM~\citep{wu2025mgmotionllm}, we report m2dt accuracy at two levels:
(i) \textbf{Sequence level}: the model emits the full multi-snippet detailed text for the whole motion as a single document, and linguistic metrics (BLEU, ROUGE, BERTScore) are computed against the ground-truth document.
(ii) \textbf{Snippet level}: the predicted and ground-truth documents are each split by $\langle \texttt{SEP} \rangle$ into per-snippet descriptions, and the metrics are computed snippet-by-snippet and then averaged. The snippet-level score is harder than the sequence level because errors in any single 0.5\,s window are penalized individually rather than smoothed over the full motion.

\paragraph{Our evaluation.}
The detailed-text format used by FineMotion is structurally different from the free-form HumanML3D captions our model was trained on in Section~\ref{sec:exp}, so we train a dedicated LoRA adapter for the m2dt task on the FineMotion training split, keeping the SMD conversion (All-26, Qwen2.5-7B base, identical hyperparameters to Section~\ref{sec:exp}) and the LoRA configuration unchanged. The SMD pipeline itself is rule-based and is not re-trained. We compare against MG-MotionLLM at its largest reported size (T5-Large, the strongest configuration in Table~4 of~\citep{wu2025mgmotionllm}), which is also trained on FineMotion (within their 28-task multi-granularity scheme), giving a fair head-to-head on the same data.

\begin{table}[h]
\centering
\caption{Fine-grained motion captioning on the FineMotion test set, evaluated under the m2dt protocol of MG-MotionLLM~\citep{wu2025mgmotionllm} at both the sequence and snippet levels. Higher is better for all metrics. MG-MotionLLM numbers are taken from their Table~4 (T5-Large variant); our model is the same Qwen2.5-7B + LoRA + All-26 SMD checkpoint as Table~\ref{tab:main}.}
\label{tab:fg_caption}
\resizebox{\textwidth}{!}{%
\begin{tabular}{llccccc}
\toprule
Level & Method & B@1$\uparrow$ & B@4$\uparrow$ & B@7$\uparrow$ & ROUGE$\uparrow$ & BERTScore$\uparrow$ \\
\midrule
\multirow{2}{*}{Sequence}
  & MG-MotionLLM (T5-Large)~\citep{wu2025mgmotionllm} & 81.73 & 66.29 & 53.56 & 65.8 & 52.4 \\
  & \textbf{Ours (SMD All-26)} & 86.95 & 69.01 & 55.28 & 66.1 & 56.1 \\
\midrule
\multirow{2}{*}{Snippet}
  & MG-MotionLLM (T5-Large)~\citep{wu2025mgmotionllm} & 68.25 & 49.26 & 37.07 & 61.4 & 52.2 \\
  & \textbf{Ours (SMD All-26)} & 69.94 & 51.33 & 40.07 & 62.6 & 56.9 \\
\bottomrule
\end{tabular}
}
\end{table}

\paragraph{Discussion.}
SMD outperforms MG-MotionLLM on every metric at both evaluation levels. At the sequence level, BLEU@1 improves by $+5.22$ points (81.73 $\to$ 86.95) and BERTScore by $+3.7$, indicating more accurate transcription of the per-body-part movement vocabulary. The gap is preserved at the harder snippet level---where errors in any single 0.5\,s window are penalized individually---with BERTScore actually growing to $+4.7$, suggesting that SMD's predictions remain semantically faithful even at fine temporal granularity. We attribute this to SMD's structural fit with the FineMotion format: SMD's per-joint, time-localized segments already provide exactly the kind of body-part-and-time-indexed information the m2dt task requires, so the LLM only has to learn to phrase it in MG-MotionLLM's snippet template, rather than recover the joint-level structure from a compressed motion latent.

\section{Experimental Setup and Costs}

\subsection{Datasets, baselines, and 10-option standardization}
\label{app:experiments_setup}

\paragraph{Original benchmarks.}
BABEL-QA~\citep{endo2023humanmotionqa} defines three question types over BABEL/AMASS motions: action queries (``What action does the person do after X?''), body part queries (``What body part does the person use?''), and direction queries (``What direction does the person move?'').
The original benchmark uses all valid labels as options, resulting in 4 options for direction, 8 for body part, and 20 for action.
HuMMan-QA~\citep{liu2024imore} follows the same format but uses the HuMMan dataset, which has a much larger label vocabulary: up to 155 action labels, 18 body parts, and 6 directions.

\paragraph{10-option standardization.}
The varying number of options (4--155) makes cross-method comparison difficult, as random-guess accuracy ranges from 25\% (4 options) to 0.6\% (155 options).
We standardize all questions to at most 10 options: for questions with more than 10 original options, we randomly sample 9 distractors plus the correct answer; for questions with fewer than 10 options, we retain the original set.
This yields an average of 8.6 options for BABEL-QA and 10.0 for HuMMan-QA, with random-guess accuracy of approximately 11.6\% and 10.0\% respectively.
The 10-option question files are generated once and shared across all methods for fair comparison.

\paragraph{Zero-shot QA answer extraction.}
The zero-shot QA results in Table~\ref{tab:zeroshot} use the same evaluation protocol as the fine-tuned runs (exact-string matching against the candidate option text) plus a lightweight regex pass that strips conversational prefixes (e.g., ``The answer is'', ``Option:'', ``A.'', leading/trailing punctuation) and lowercases the prediction before comparison. This pass is applied uniformly to every method's outputs. The regex normalization is necessary because base LLMs without task-specific fine-tuning often wrap the answer in a sentence (e.g., ``The correct option is left hand.''); without it, exact-string matching would penalize formatting habits rather than reasoning ability. The same regex is reused in the open-vocabulary evaluation of Appendix~\ref{app:openvocab}.

\paragraph{Baseline retraining.}
All QA baselines reported in Table~\ref{tab:main} expect MotionLLM~\citep{chen2024motionllm} are retrained and evaluated on the standardized 10-option format:

IMoRe~\citep{liu2024imore} is a specialized motion QA model that uses classifier heads over a fixed label vocabulary.
We retrain IMoRe on the 10-option training data for both BABEL-QA and HuMMan-QA.
At evaluation, we extract the full logit vector from the classifier, mask it to only the 10 candidate options for each question, and take the argmax over the masked logits.
This ensures the model selects from exactly the same option set as all other methods.

MotionLLM~\citep{chen2024motionllm} results are taken from the original paper, as neither the training data nor the checkpoints are publicly available for retraining.
The reported 43.6\% on BABEL-QA uses the original option format and is therefore not directly comparable to our 10-option results, though we include it as a reference.
MotionLLM uses Vicuna-13B as backbone with a learned motion encoder that projects SMPL motion features into the LLM's embedding space via a linear projection layer, trained with LoRA on a combined video-motion dataset (MoVid).

NSPose~\citep{endo2023humanmotionqa} is the original motion QA method that defines the task and proposes a neuro-symbolic framework.
It uses a pretrained Motion ViT encoder and executes modular programs recursively over the learned motion representations.
We retrain NSPose on the 10-option format using their released code.

\paragraph{Captioning baselines.}
For the captioning experiments in Table~\ref{tab:main}, baseline numbers for TM2T, MotionGPT, LaMP, MoTe, and MotionGPT3 are taken from the MotionGPT3 paper~\citep{chen2024motiongpt3}, which evaluates all methods under the same protocol on HumanML3D, except for CIDEr which we re-compute on the released checkpoints with the cleaned \texttt{pycocoevalcap} pipeline described in Appendix~\ref{app:metrics} (the original numbers were inflated by extraneous symbols such as colons and newlines in the predictions).
MG-MotionLLM~\citep{wu2025mgmotionllm} numbers are from their CVPR 2025 paper.

\paragraph{MotionGPT3-Qwen.}
To provide a controlled comparison isolating the effect of motion representation, we replicate the MotionGPT3 paradigm on the same Qwen2.5-7B backbone used by our method.
Specifically, we use the pretrained MotionGPT3 VAE encoder (frozen) to encode each motion into a continuous latent vector of dimension 256.
This latent is projected to $N$ LLM-sized token embeddings via a two-layer MLP (256 $\to$ 2048 $\to$ $N \times 3584$, where 3584 is the Qwen2.5-7B hidden dimension).
Training follows the two-stage LLaVA~\citep{liu2024llava} protocol: in stage 1, the LLM is frozen and only the projection MLP is trained; in stage 2, LoRA is added to the LLM and both the MLP and LoRA are trained jointly.
Both stages use learning rate 1e-4 (stage 1) and 1e-4 (stage 2), with 5 epochs each.
For QA, the motion tokens are prepended to the question in the prompt; for captioning, they replace the SMD text.
We evaluate configurations with $N \in \{4, 32, 64, 128\}$ projection tokens (see Appendix~\ref{app:encoder_sweep}).

\subsection{Evaluation metric definitions}
\label{app:metrics}

We evaluate motion-to-text captioning using two categories of metrics.

\paragraph{Text-motion alignment metrics.}
These metrics assess whether the generated caption semantically matches the corresponding motion, using the pretrained T2M evaluator~\citep{guo2022t2m} which provides a shared embedding space for text and motion.

R-Precision (R@$k$, $k \in \{1,2,3\}$) measures retrieval accuracy: given a pool of 32 motions, the generated caption is used to retrieve the best-matching motion via cosine similarity in the T2M embedding space.
R@$k$ reports the fraction of times the ground-truth motion appears in the top-$k$ retrieved results, averaged over all test samples.
Higher R-Precision indicates better semantic alignment between the generated text and the original motion.

MM-Distance (Multi-Modal Distance) computes the average Euclidean distance between the T2M embeddings of each generated caption and its ground-truth motion.
Lower MM-Distance indicates that generated captions are semantically closer to their corresponding motions in the learned embedding space.

Both metrics are computed using the Comp\_v6\_KLD01 checkpoint of the T2M evaluator.

\paragraph{Linguistic metrics.}
These metrics evaluate the textual quality of generated captions by comparing them against reference captions using standard NLP evaluation tools.

BLEU@$k$~\citep{papineni2002bleu} measures modified $n$-gram precision between the generated and reference captions, with $k$ indicating the maximum $n$-gram order.
We report BLEU@1 (unigram precision, capturing word overlap) and BLEU@4 (up to 4-gram precision, capturing phrase-level similarity).

ROUGE-L~\citep{lin2004rouge} computes the longest common subsequence (LCS) between the generated and reference captions, measuring recall-oriented overlap at the sentence level.

CIDEr~\citep{vedantam2015cider} uses TF-IDF weighted $n$-gram matching to evaluate consensus between the generated caption and multiple reference captions, designed specifically for image/video captioning evaluation.

BERTScore~\citep{zhang2019bertscore} computes token-level cosine similarity between contextual embeddings of the generated and reference captions using a pretrained BERT model, capturing semantic similarity beyond surface-level $n$-gram overlap.

All linguistic metrics are computed on motions with 20--200 frames using pycocoevalcap, matching the data range used by MotionGPT3~\citep{chen2024motiongpt3}.

\paragraph{CIDEr implementation.}
Since the original CIDEr computation in prior works inadvertently includes extraneous symbols such as colons and newlines, which affect the scores. We compute CIDEr using the \texttt{pycocoevalcap} library (the standard COCO Caption evaluation toolkit), which implements CIDEr-D from Vedantam et al.~\citep{vedantam2015cider}.
The implementation uses $n$-grams up to $n{=}4$ with a Gaussian length-penalty standard deviation of $\sigma{=}6.0$, and computes IDF statistics directly from the reference captions of the evaluation subset.
Following standard practice, when a motion has multiple reference captions we treat them as a set of references for the same query, rather than averaging individual scores.
We note that CIDEr is highly sensitive to corpus-level IDF statistics and to caption tokenization; minor differences in the evaluation subset (e.g., length-filtered vs.\ full test set) or in the underlying tokenizer can shift the absolute CIDEr value by several points without affecting the relative ordering of methods.
For this reason, we report all baseline and SMD CIDEr numbers computed with the same \texttt{pycocoevalcap} pipeline on the identical evaluation subset, ensuring a fair comparison even if the absolute values differ from those reported in the original papers.

\subsection{Training and inference cost comparison}
\label{app:cost}

This appendix details the training and inference costs of SMD relative to encoder-based motion LLMs summarized in Section~\ref{sec:backbone}. Table~\ref{tab:cost} lists the head-to-head comparison.

\paragraph{Training cost.}
Retraining SMD on a new LLM backbone takes 2--8 GPU-hours for QA and 6--20 hours for captioning on a single H200, with the captioning range varying by backbone size (3B--14B parameters) and SMD variant (Top-3 $\approx$1k tokens vs All-26 $\approx$4k tokens); the $\sim$20 GPU-hour figure quoted in Section~\ref{sec:exp} corresponds to the All-26 + Qwen2.5-7B reference run reported in Table~\ref{tab:main}. Only $\sim$40M LoRA parameters are updated while the base LLM is frozen and the SMD conversion itself is rule-based and has no trainable parameters. In contrast, encoder-based methods such as MotionGPT3 and MG-MotionLLM retrain the full pipeline on each new LLM, including the motion encoder and the alignment module. This involves hundreds of millions of parameters and a multi-stage schedule (pretrain the motion encoder, then the projection or alignment module, then the LLM), on the order of tens of GPU-hours even for their relatively small backbones.

\paragraph{Inference cost.}
The main trade-off of SMD is a longer input sequence. Encoder-based methods compress a motion into $\sim$256 tokens, whereas SMD produces $\sim$1{,}000 tokens for the Top-3 variant and $\sim$4{,}000 tokens for the All-26 variant. We reported the cost comparision in Table~\ref{tab:cost}. On a single H200 GPU, the average latency of our approach for captioning task is 915 ms/sample (1.1 samples/s) for Top-3 SMD and 1{,}154 ms/sample (0.9 samples/s) for All-26 SMD, using approximately 15.5 GB of GPU memory. To isolate the input-length effect from the backbone size, we also benchmark the same-backbone MotionGPT3-Qwen baseline (Qwen2.5-7B + LoRA + frozen MotionGPT3 VAE + MLP projection, $N{=}4$ motion tokens) under the identical protocol; it averages $300$ ms/sample ($3.3$ samples/s) at $14.7$\,GB GPU memory.

\begin{table}[h]
\centering
\caption{Comparison of training and inference costs across LLM backbones. Our method retrains only a lightweight LoRA adapter, whereas prior methods must retrain the full alignment pipeline.}
\label{tab:cost}
\resizebox{0.75\textwidth}{!}{%
\begin{tabular}{lccc}
\toprule
& Ours (SMD) & MotionGPT3 & MG-MotionLLM \\
\midrule
LLM backbone & Any (8 tested) & GPT-2 (124M) & T5-Base (220M) \\
Motion representation & Rule-based text & Learned VAE & Learned VQ-VAE \\
Training stages & 1 (LoRA only) & 3 & 2 \\
Retrain on new LLM & LoRA only & Full pipeline & Full pipeline \\
Trainable params & $\sim$40M & $\sim$124M & $\sim$220M \\
Caption train time & $\sim$6--20 h & $\sim$60 h$^*$ & $\sim$60 h$^*$ \\
Input tokens (avg) & $\sim$1{,}000--4{,}000 & $\sim$256 & $\sim$256 \\
Inference (ms/sample) & 915--1{,}154 & \multicolumn{2}{c}{$<$100 (smaller LLM)} \\
\bottomrule
\multicolumn{4}{l}{\footnotesize $^*$Estimated total multi-stage training time including encoder pretraining.}
\end{tabular}
}
\end{table}

\subsection{MotionGPT3-Qwen: projection token sweep}
\label{app:encoder_sweep}

Table~\ref{tab:encoder_sweep} reports the full token sweep for MotionGPT3-Qwen on both QA and captioning.
As the number of projection tokens increases from 4 to 128, the projection MLP grows from 15M to 940M parameters.
For QA, all configurations achieve similar BABEL-QA accuracy (47--50\%), far below SMD (66.7\%), and perform poorly on HuMMan-QA (15--24\% vs.\ SMD 90.1\%) due to the cross-domain issue discussed in Section~\ref{sec:exp} and Appendix~\ref{app:feature_stats}.
For captioning, performance holds steady at 4 and 32 tokens but degrades with larger MLPs (64 and 128 tokens), as the projection layer overfits on the limited 14{,}148 training pairs.

\begin{table}[h]
\centering
\caption{MotionGPT3-Qwen with varying projection tokens. All configurations use the same frozen VAE and two-stage training.}
\label{tab:encoder_sweep}
\resizebox{\textwidth}{!}{%
\begin{tabular}{lrccccccccccc}
\toprule
Tokens & Params & BABEL-QA & HuMMan-QA & R@1$\uparrow$ & R@2$\uparrow$ & R@3$\uparrow$ & MMD$\downarrow$ & B@1$\uparrow$ & B@4$\uparrow$ & R-L$\uparrow$ & CIDEr$\uparrow$ & BS$\uparrow$ \\
\midrule
4 & 30M & 50.1 & 22.0 & .555 & .753 & .843 & 2.47 & 59.43 & 19.60 & 45.82 & 46.13 & 42.87 \\
32 & 235M & 50.4 & 23.5 & .570 & .759 & .846 & 2.44 & 57.22 & 17.53 & 44.31 & 43.52 & 41.76 \\
64 & 470M & 49.9 & 14.6 & .547 & .739 & .826 & 2.60 & 56.09 & 17.25 & 42.38 & 37.89 & 37.86 \\
128 & 940M & 47.3 & 23.6 & .500 & .710 & .803 & 2.65 & 59.33 & 19.78 & 44.73 & 41.68 & 40.74 \\
\midrule
\multicolumn{2}{l}{\textbf{SMD (ours)}} & \textbf{66.7} & \textbf{90.1} & \textbf{.584} & \textbf{.794} & \textbf{.883} & \textbf{2.35} & \textbf{63.45} & \textbf{22.67} & \textbf{47.80} & \textbf{53.16} & \textbf{45.58} \\
\bottomrule
\end{tabular}
}
\end{table}


\section{Qualitative Analysis}
\label{app:qualitative}

This section provides side-by-side qualitative comparisons of our SMD-based approach against the baselines on motion QA and motion captioning.

\subsection{Motion QA}
\label{app:qualitative_qa}

\begin{figure}[h]
\centering
\includegraphics[width=\textwidth]{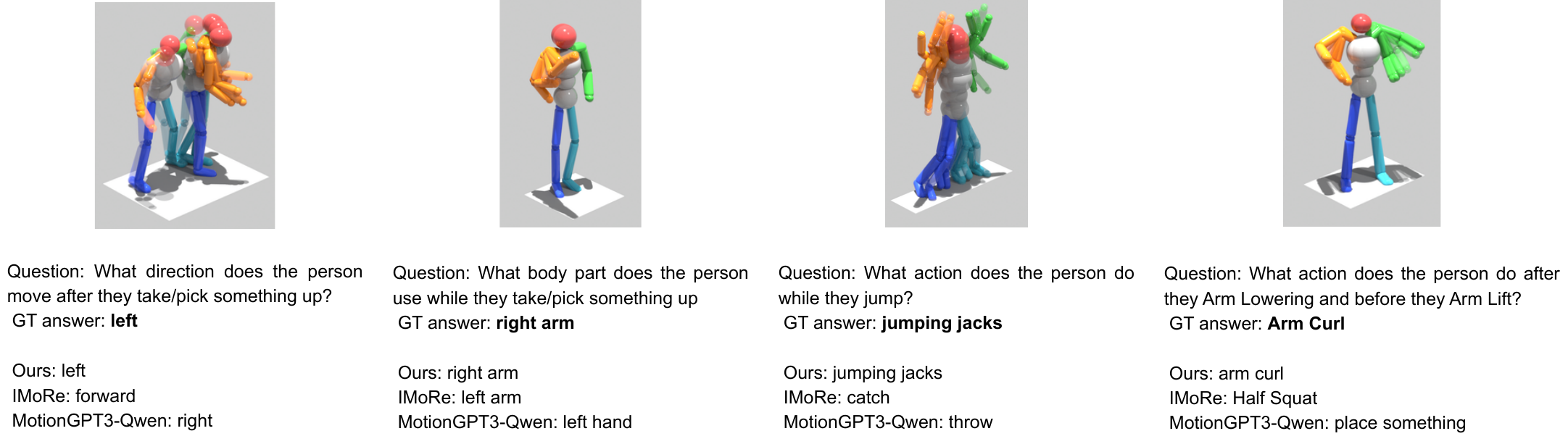}
\caption{Qualitative motion QA comparison on BABEL-QA and HuMMan-QA. Each panel shows a representative frame of the motion, the question, the ground-truth answer, and the predictions from IMoRe~\citep{liu2024imore}, MotionGPT3-Qwen, and our SMD-based approach. SMD answers all four examples correctly; both baselines confuse the discriminative body part, direction, or action category in every case.}
\label{fig:qual_qa}
\end{figure}

Figure~\ref{fig:qual_qa} shows four motion QA examples spanning the three query types (direction, body part, action) and both datasets (BABEL-QA, HuMMan-QA). On all four cases SMD picks the correct answer while both IMoRe and MotionGPT3-Qwen pick a wrong but semantically related option---e.g., \textit{forward} or \textit{right} instead of \textit{left}, \textit{left arm/hand} instead of \textit{right arm}, \textit{catch} or \textit{throw} instead of \textit{jumping jacks}, and \textit{Half Squat} or \textit{place something} instead of \textit{Arm Curl}. The grounded, body-part-indexed text that SMD provides makes it easy for the LLM to attend to the discriminative joint.

\subsection{Motion captioning}
\label{app:qualitative_caption}

\begin{figure}[h]
\centering
\includegraphics[width=\textwidth]{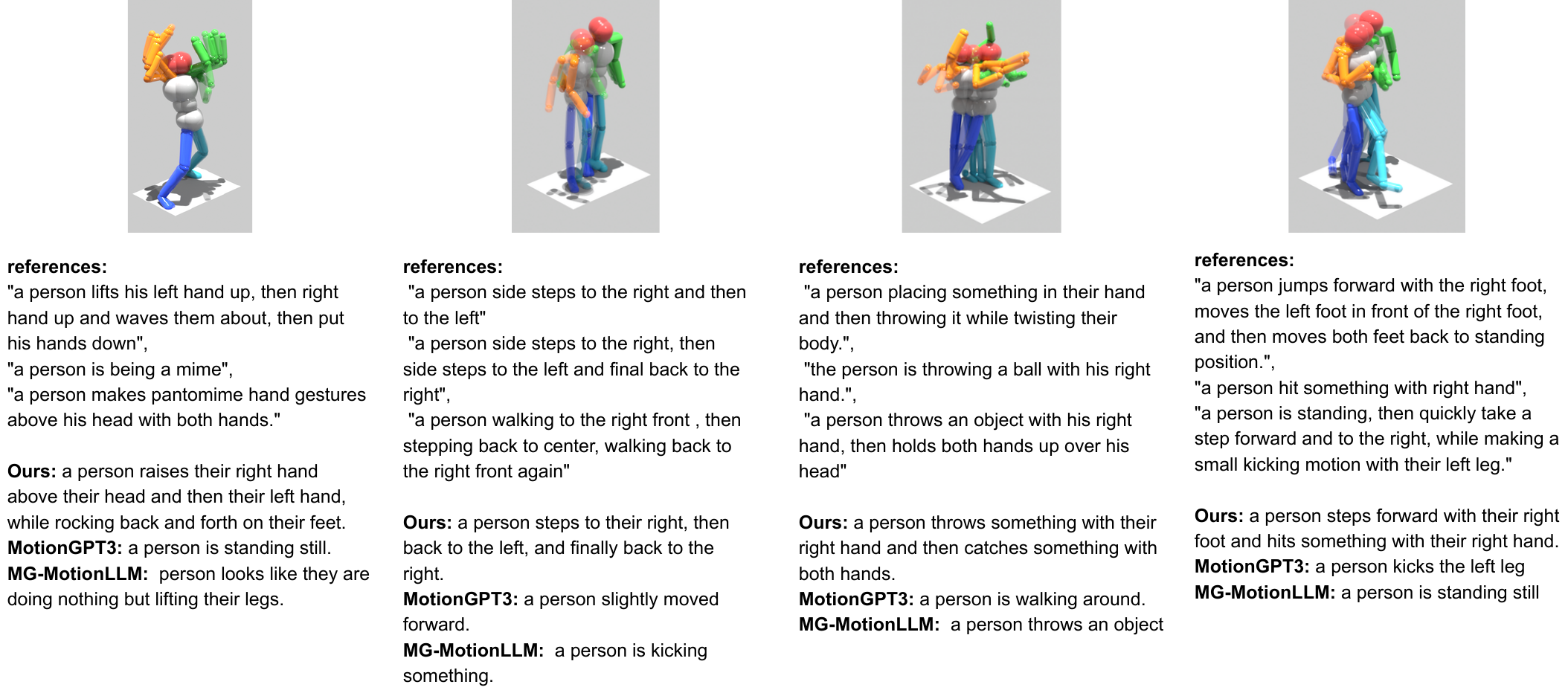}
\caption{Qualitative motion captioning comparison on HumanML3D. Each panel shows a representative motion frame, the ground-truth reference captions, and the captions produced by MotionGPT3~\citep{chen2024motiongpt3}, MG-MotionLLM~\citep{wu2025mgmotionllm}, and our SMD-based approach. SMD captures the multi-step, body-part-specific structure of the references on all four examples; the baselines either give generic descriptions, swap to a visually similar action, or recover only the coarse action label without the discriminative body part or temporal sequence.}
\label{fig:qual_caption}
\end{figure}

Figure~\ref{fig:qual_caption} shows four motion captioning examples from the HumanML3D test set comparing our SMD model against MotionGPT3~\citep{chen2024motiongpt3} and MG-MotionLLM~\citep{wu2025mgmotionllm}. On all four examples SMD produces detailed, body-part-grounded captions that reflect the multi-step structure of the references---e.g., \textit{``raises their right hand above their head and then their left hand, while rocking back and forth''} for the pantomime gesture, \textit{``steps to their right, then back to the left, and finally back to the right''} for the side-step sequence, and \textit{``throws something with their right hand and then catches something with both hands''} for the throwing motion. The two baselines, lacking this body-part trace, often fall back on generic descriptions (\textit{``standing still''}, \textit{``slightly moved forward''}, \textit{``walking around''}), replace the action with a visually similar one (\textit{``kicking something''}, \textit{``kicks the left leg''}), or recover only the coarse action label without its discriminative body part or temporal structure (\textit{``throws an object''}).

\section{Broader Impact}

\label{app:broader_impact}

Accurate human motion understanding in natural language can benefit a range of applications, such as virtual coaches, rehabilitation assistants, and accessible motion descriptions for visually impaired users. Because our Structured Motion Description is grounded in biomechanical joint angles already used in clinical gait analysis and sports science, it is particularly suited to settings where interpretable, body-part-level feedback matters, including physical therapy progress tracking, sports coaching, and dance or martial arts instruction. Representing motion in a structured textual form may also help digitally preserve movement-based cultural heritage such as folk dances and traditional martial arts, complementing video records with human-readable descriptions. The other side of the coin is privacy: because motion and gait patterns can be used to re-identify individuals, care needs to be taken to avoid applications that monitor specific groups of people without consent.

\end{document}